\documentclass[Afour,sageh,times]{sagej}
\usepackage{etex} 

\makeatletter
\def\gnewcommand{\g@star@or@long\new@command}
\def\grenewcommand{\g@star@or@long\renew@command}
\def\g@star@or@long#1{%
  \@ifstar{\let\l@ngrel@x\global#1}{\def\l@ngrel@x{\long\global}#1}}
\makeatother

\providecommand{\provideabbreviation}[2]{\providecommand{#1}{#2\xspace}}

\providecommand{\newabbreviation}[2]{#1\ (\grenewcommand{#1}{#2\xspace}#1)}

\newcommand{\LfD    }{Learning from Demonstration\xspace}

\provideabbreviation{\TP     }{task-parametrised}
\provideabbreviation{\GMM    }{Gaussian mixture model}
\provideabbreviation{\GMR    }{Gaussian mixture regression}

\newcommand{\HRI    }{human robot interaction\xspace}

\newcommand{\TPGMM	}{Task-Parameterised Gaussian Mixture Model\xspace}

\newcommand{\EICI	}{NF\xspace}
\newcommand{\EICII	}{VF\xspace}
\newcommand{\EICIII	}{VR\xspace}

\newcommand{\EIICI	}{NF\xspace}
\newcommand{\EIICII	}{RF\xspace}
\newcommand{\EIICIII	}{BF\xspace}
\newcommand{\EIICIV	}{SF\xspace}

\mathchardef\mhyphen="2D   

\newcommand{\T}     {\top}                

\newcommand{\estimated} [1]{\tilde{#1}}

\newcommand{\nd}      {j}                              
\newcommand{\nk}      {k}                              
\newcommand{\Nk}      {\mathcal{\MakeUppercase{\nk}}}  

\renewcommand{\nu}  {q}                   

\newcommand{\Is} {\mathcal{I}} 
\newcommand{\policy} {\pi} 

\newcommand{\Ts} {\beta} 
\newcommand{\spaceX} {\boldsymbol{\mathcal{X}}} 

\newcommand{\Ns} {n} 

\newcommand{\statept} {\boldsymbol{\xi}} 

\newcommand{\spaceR} {\boldsymbol{\mathfrak{R}}}  
\newcommand{\spacer} {\boldsymbol{\mathfrak{r}}}  
\newcommand{\spaceIh}{\boldsymbol{\mathfrak{I}}_{h}}  %
\newcommand{\spaceB} {\boldsymbol{\mathfrak{B}}}  
\newcommand{\spaceb} {\boldsymbol{\mathfrak{b}}}  
\newcommand{\spacePi} {\boldsymbol{\Pi}}  
\newcommand{\spaceM} {\boldsymbol{\mathfrak{M}}}  

\newcommand{\Hfact} {\mathcal{Q}}  

\newcommand{\traj} {\mathcal{T}}
\newcommand{\efficacy} {\varepsilon}
\newcommand{\efficiency} {\eta}

\providecommand{\figurename}{Fig.}
\providecommand*{\sref}[1]{\S\ref{s:#1}}            
\providecommand*{\tref}[1]{\tablename~\ref{t:#1}}   
\providecommand*{\fref}[1]{\figurename~\ref{f:#1}}  
\providecommand*{\eref}[1]{(\ref{e:#1})}            

\usepackage[inline]{enumitem}
\setlist{nolistsep}
\providecommand{\il}[1]{\begin{enumerate*}[label=(\roman*)]#1\end{enumerate*}}

\providecommand{\eg}{\textit{e.g.,}~} %
\providecommand{\ie}{\textit{i.e.,}~} %
\providecommand{\etc}{\textit{etc.}\xspace}  %
\providecommand{\naive}{na\"ive\xspace}      %
\newlist{hypotheses}{enumerate}{10}
\setlist[hypotheses]{label*=$\mathbf{H_{\arabic*}}$:,
                     ref=$\mathbf{H_{\arabic*}}$}

\renewcommand{\cite}[1]{\citep{#1}}
\setcitestyle{aysep={,}} 

\usepackage{color}
\usepackage{textcomp}
\usepackage[percent]{overpic}
\usepackage{wrapfig}
\usepackage{times}
\usepackage{graphicx}\graphicspath{{fig/}} 
\usepackage{epstopdf}
\usepackage{algorithm}
\usepackage{algorithmic}
\usepackage{multirow} 
\usepackage{multicol}
\usepackage{array}
\usepackage{mdwmath}
\usepackage{mdwtab}
\usepackage{rotating}
\usepackage[list=off,font=footnotesize]{caption}
\usepackage{amssymb}
\usepackage{verbatim}
\usepackage[labelfont=normalfont]{subcaption}
\usepackage{pdfpages}
\usepackage{transparent}
\usepackage{xspace}
\usepackage{color,soul} 
\usepackage{url}
\usepackage{textcomp}
\usepackage{enumitem}
\usepackage{booktabs}
\usepackage{balance}
\usepackage{tabularx}
\usepackage{pdfpages}
\usepackage{pdflscape}
\usepackage{flushend}
\usepackage{moreverb,url}

\usepackage[draft,colorlinks,bookmarksopen,bookmarksnumbered,citecolor=red,urlcolor=red]{hyperref}

\makeatletter\newcommand{\manuallabel}[2]{\def\@currentlabel{#2}\label{#1}}\makeatother

\newcommand\BibTeX{{\rmfamily B\kern-.05em \textsc{i\kern-.025em b}\kern-.08em
\kern-.1667em\lower.7ex\hbox{E}\kern-.125emX}}

\def\volumeyear{2019}

\begin{document}

\runninghead{Sena and Howard}

\title{Quantifying Teaching Behaviour in Robot Learning from Demonstration}

\author{Aran Sena\affilnum{1} and Matthew Howard\affilnum{1}}

\affiliation{\affilnum{1}Department of Informatics, King's College London, UK}

\corrauth{Aran Sena, King's College London
Strand,
London,
WC2R 2LS, UK.}

\email{aran.sena@kcl.ac.uk}

\twocolumn

\begin{abstract}
Learning from demonstration allows for rapid deployment of robot manipulators to a great many tasks, by relying on a person showing the robot what to do rather than programming it. While this approach provides many opportunities, measuring, evaluating and improving the person's teaching ability has remained largely unexplored in robot manipulation research. To this end, a model for learning from demonstration is presented here which incorporates the teacher's understanding of, and influence on, the learner. The proposed model is used to clarify the teacher's objectives during learning from demonstration, providing new views on how teaching failures and efficiency can be defined. The benefit of this approach is shown in two experiments ($\Ns=30$ and $\Ns=36$, respectively), which highlight the difficulty teachers have in providing effective demonstrations, and show how $\sim169-180\%$ improvement in teaching efficiency can be achieved through evaluation and feedback shaped by the proposed framework, relative to unguided teaching.

\end{abstract}

\keywords{Learning from Demonstration, Machine Teaching, Human Robot Interaction}

\maketitle
%
\section{Introduction}\label{s:introduction}
The rise of collaborative robots---robots designed to work safely in close proximity with people---is making it more common for people interacting with robots to have little or no technical background. These systems reduce the requirement for conventional programming skills for robot operation through the use of simplified programming interfaces, making deployment faster and more flexible. Of particular interest for deploying such systems is \newabbreviation{\LfD}{LfD} as a way to enable \textit{novice} users (\ie people with no relevant technical background) to \emph{teach} robots to perform useful work, similarly to how they would train an ordinary co-worker \cite{Billard2008, Argall2009, Calinon2017LearningControl}.

\begin{figure}
     \def\svgwidth{\linewidth}
\begingroup%
  \makeatletter%
  \providecommand\color[2][]{%
    \errmessage{(Inkscape) Color is used for the text in Inkscape, but the package 'color.sty' is not loaded}%
    \renewcommand\color[2][]{}%
  }%
  \providecommand\transparent[1]{%
    \errmessage{(Inkscape) Transparency is used (non-zero) for the text in Inkscape, but the package 'transparent.sty' is not loaded}%
    \renewcommand\transparent[1]{}%
  }%
  \providecommand\rotatebox[2]{#2}%
  \newcommand*\fsize{\dimexpr\f@size pt\relax}%
  \newcommand*\lineheight[1]{\fontsize{\fsize}{#1\fsize}\selectfont}%
  \ifx\svgwidth\undefined%
    \setlength{\unitlength}{226.46697758bp}%
    \ifx\svgscale\undefined%
      \relax%
    \else%
      \setlength{\unitlength}{\unitlength * \real{\svgscale}}%
    \fi%
  \else%
    \setlength{\unitlength}{\svgwidth}%
  \fi%
  \global\let\svgwidth\undefined%
  \global\let\svgscale\undefined%
  \makeatother%
  \begin{picture}(1,0.50352938)%
    \lineheight{1}%
    \setlength\tabcolsep{0pt}%
    \put(0,0){\includegraphics[width=\unitlength,page=1]{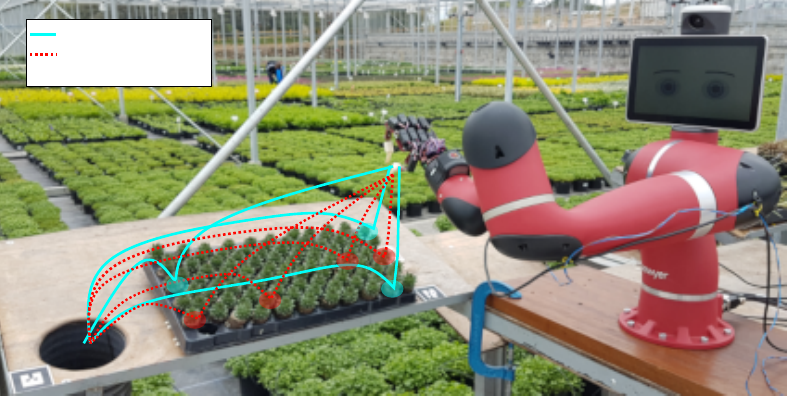}}%
    \put(0.07514611,0.45178784){\color[rgb]{0,0,0}\makebox(0,0)[lt]{\lineheight{1.25}\smash{\begin{tabular}[t]{l}\scriptsize{Demonstrations}\end{tabular}}}}%
    \put(0.07514611,0.42691388){\color[rgb]{0,0,0}\makebox(0,0)[lt]{\lineheight{1.25}\smash{\begin{tabular}[t]{l}\scriptsize{Generated}\end{tabular}}}}%
    \put(0,0){\includegraphics[width=\unitlength,page=2]{demos_repos_setting.pdf}}%
    \put(0.07514611,0.40121216){\color[rgb]{0,0,0}\makebox(0,0)[lt]{\lineheight{1.25}\smash{\begin{tabular}[t]{l}\scriptsize{Target}\end{tabular}}}}%
  \end{picture}%
\endgroup%

		\caption{Learning from Demonstration can be used to learn a skill---here, disposal of unhealthy plants from a plant tray---from a limited set of user-provided demonstrations. Key to this is the ability to generalise from the demonstrated trajectories (cyan) to determine the appropriate trajectory to pick plants from previously unseen locations (red).}%
		\label{f:gex}%
\end{figure}

\begin{figure*}
	 \def\svgwidth{\linewidth}
\begingroup%
  \makeatletter%
  \providecommand\color[2][]{%
    \errmessage{(Inkscape) Color is used for the text in Inkscape, but the package 'color.sty' is not loaded}%
    \renewcommand\color[2][]{}%
  }%
  \providecommand\transparent[1]{%
    \errmessage{(Inkscape) Transparency is used (non-zero) for the text in Inkscape, but the package 'transparent.sty' is not loaded}%
    \renewcommand\transparent[1]{}%
  }%
  \providecommand\rotatebox[2]{#2}%
  \newcommand*\fsize{\dimexpr\f@size pt\relax}%
  \newcommand*\lineheight[1]{\fontsize{\fsize}{#1\fsize}\selectfont}%
  \ifx\svgwidth\undefined%
    \setlength{\unitlength}{464.88188976bp}%
    \ifx\svgscale\undefined%
      \relax%
    \else%
      \setlength{\unitlength}{\unitlength * \real{\svgscale}}%
    \fi%
  \else%
    \setlength{\unitlength}{\svgwidth}%
  \fi%
  \global\let\svgwidth\undefined%
  \global\let\svgscale\undefined%
  \makeatother%
  \begin{picture}(1,0.22560976)%
    \lineheight{1}%
    \setlength\tabcolsep{0pt}%
    \put(0,0){\includegraphics[width=\unitlength,page=1]{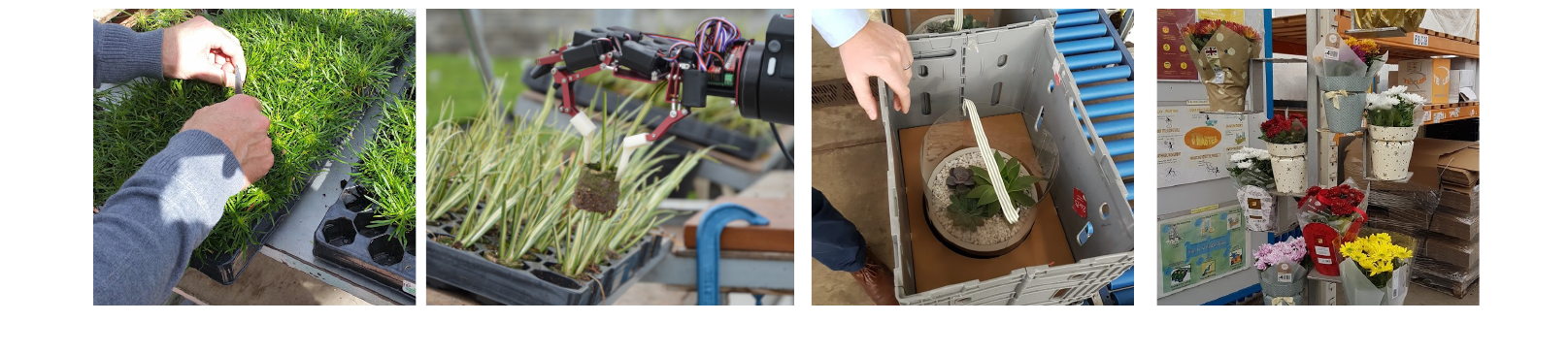}}%
    \put(0.26998055,0.00492467){\color[rgb]{0,0,0}\makebox(0,0)[t]{\lineheight{1.25}\smash{\begin{tabular}[t]{c}{(a)}\end{tabular}}}}%
    \put(0.73176924,0.00492449){\color[rgb]{0,0,0}\makebox(0,0)[t]{\lineheight{1.25}\smash{\begin{tabular}[t]{c}{(b)}\end{tabular}}}}%
  \end{picture}%
\endgroup%

	 \vspace{-0.6cm}
	 \caption{``Low-batch'' manual tasks found in the ornamental horticulture industry. (a) Quality assessment automation is often hampered due to companies growing a large variety of products, and (b)  packaging often requires manual labour due to varying retail packaging requirements that can change seasonally, thus creating a need for more flexible automation systems.}      
	 \label{f:robot}
\end{figure*}

Key to making \LfD practical in the real world, is ensuring efficient skill-learning by the robot. Efficient skill-learning in this context means the robot is able to not just learn to perform the demonstrated task \textit{as it was shown by the person}, but to \emph{generalise beyond the specific demonstrations provided}, to new circumstances. This ability helps the robot to adapt to uncertainty during operation and reduces the teaching effort for the person doing the teaching. 

Deploying systems that can learn generalised tasks from human teachers presents a number of challenges for both the learner and the teacher. When a system incorporates human input into its learning process, the system performance strongly depends on the quality of the data provided. As noted by \citet{Argall2009}, key issues that can arise as a result of poor teaching in \LfD include \il{%
	\item \emph{undemonstrated states}, 
	\item \emph{ambiguous demonstrations}, and
	\item \emph{unsuccessful demonstrations}
}, all of which might confuse the robot learner, resulting in poor performance during skill realisation.
Poor teaching is especially prevalent with novice teachers and can be attributed to teachers lacking a shared mental model with the robot, or a lack of understanding of \emph{how} and \emph{what} the robot learns during the teaching process \cite{Cakmak2014,Hellstrom2018UnderstandableHow}.

The \emph{role of the teacher} in \LfD systems is an under-explored area of research, with a lack of formal methods for \emph{measuring, evaluating and modifying teaching behaviour}. To address this, a new theoretical framework for the assessment and guidance of \LfD is presented here. Built on prior work in machine teaching, active learning, and \LfD, the proposed framework enables a formal analysis of human-robot teaching during \LfD to help people more effectively teach robots new skills. It facilitates \il{\item the definition of metrics to assess users' quality of teaching, \item the quantitative measurement of teaching failures, such as undemonstrated states and ambiguous demonstrations, and \item aids the development of tools (visualisations, procedures, \etc) that can provide a clearer view of how to guide the teaching process}. 

The effectiveness of this framework as a means of helping novices during \LfD is then shown in two experiments with novice users, extending the work presented in \cite{Sena2018TeachingLearners}. It is shown in these validation experiments that the evaluation and feedback tools derived from the proposed framework allow for improving teaching performance, in the order of a $180\%$ improvement relative to unguided conditions. This highlights the benefit of measuring, evaluating, and modifying \emph{human teaching practice} during \LfD, and provides insight into how \LfD could be more widely used as a practical tool for deploying robotic automation.

\section{Background and Related Work}\label{s:related}
Presented here is background on the field of \LfD, and the current gap in considering the role of the teacher. Insights from the related domain of algorithmic machine teaching are then considered, to point to new directions for \LfD modelling.

Learning from demonstration in robotics is often described as a form of supervised learning, where a teacher provides examples of a target task they would like a robot to perform, which the robot uses to learn a control policy \cite{Billard2008, Argall2009, Billing2010}. \LfD is a useful approach for robot task learning as it reduces the complexity of search spaces for learning real-world tasks, and reduces the amount of tedious programming required which helps \textit{novice users} use robot systems \cite{Billard2008, Chernova2014}.

There is a large body of research available on the policy learning aspect of \LfD, including symbolic reasoning methods \cite{Billard2008, Ahmadzadeh2015}, reinforcement learning based methods \cite{Schaal1996LearningDemonstration, Ng2000AlgorithmsLearning, Abbeel2004ApprenticeshipLearning, Kormushev2013}, dynamical system modelling methods \cite{Schaal2006DynamicRobotics, Pervez2017LearningGMMs}, probabilistic methods \cite{Asfour2006, Calinon2007b, Cederborg2010, Calinon2015, Maeda2017ActivePrimitives,Huang2018GeneralizedLearning}, particle based approaches \cite{Groth2014One-shotSimulation,Orendt2016RobotProgramming}, and geometric based methods \cite{Ahmadzadeh2018Trajectory-BasedCylinders}. There is, however, less research found in robot \LfD on the teaching provided by people to robots, particularly for learning grasping tasks. 

While expert-level knowledge may not be required with \LfD, with many possible policy learning methods, there will be many different requirements for the data provided to the learning system. The optimal teaching strategy a user should use may therefore be difficult to identify for novice users.

\subsection{Teaching Feedback and Evaluation}
Feedback from robot learners to human teachers in \LfD is often considered in pragmatic terms, with solutions determined based on the task at hand, or simply assuming the teacher will be able to successfully interpret the learner's actions and adjust their teaching behaviour accordingly. Incremental methods in \LfD are viewed as a way of gradually learning a skill, and can be adapted to help improve and/or overcome a users physical skill deficiencies \cite{Calinon2007IncrementalRobot,Hoyos2016IncrementalModel,Tykal2016IncrementallyDemonstration}. However, even though the process is iterative, there has been little research on how novice users interpret and adapt to the learner over consecutive teaching steps. 

\citet{Calinon2007IncrementalRobot} and \citet{Weiss2009TeachingHOAP-3} both describe an incremental learning system where the active role that teachers can play during the learning process is considered. In deciding where to provide a new demonstration, the authors describe the teacher observing the robot attempt the taught skill in new locations to determine how to provide the next demonstration. This overlooks the question of \textit{how} the attempts should be selected to help inform the teacher, whether the teacher will be able to effectively test the learner's knowledge, and whether the teacher will be able to correctly interpret the attempts.

Effective feedback of what a robot has learned to do can allow a teacher to provide more effective instruction to the learner, without the need for understanding how learning is taking place. This is shown in \citet{Nicolescu2003NaturalLearning}, where teachers observe robots executing learned plans. Similarly,
\citet{Argall2007LearningTeacher} describe a system for allowing a teacher to provide specific feedback on trajectories generated by a learner after teaching. This highlights the benefit of informed teaching from a human teacher, made possible through effective feedback of what the robot has learned to do.

\citet{Toris2012} presents a user study comparing three \LfD policy learning methods on a sweeping task. A common feedback point raised by participants was the need for a better understanding of what the robot is ``thinking''. While direct comparison of learning methods for usability is valuable, it is likely that specific methods will be of benefit to specific types of tasks.

An additional benefit to improving feedback in \LfD systems, for the benefit of helping the teacher to understand the learner's capabilities, is this helps to facilitate trust between the human teacher and robot learner \cite{Yang2017EvaluatingAutomation,Lewis2018TheInteraction}, though this is not directly explored in this work.

Other approaches to achieving skill learning and generalisation with minimal teaching effort have considered \textit{active learning} based approaches, where the learner decides when a new demonstration is required. This can be achieved by either randomly sampling trajectories and asking the teacher if they are acceptable, or requesting new demonstration when entering regions of high uncertainty \cite{Argall2009,Cakmak2011ActiveDemonstration,Maeda2017ActivePrimitives}. While this may be a complimentary approach to improving teaching, this does not supersede the benefit of a good teacher, as discussed in \sref{machine_teaching}, and thus the need to understand how to measure and improve teaching remains important.

Similar to issues found in \LfD feedback, evaluation of \LfD systems has tended to focus on robot learners. Evaluation of teachers has more focused on their physical skill when executing the task, such as in \citet{Ureche2015} where metrics are proposed for determining quality of bimanual demonstrations, and \citet{Cho2013} where an approach for identifying good teaching is determined by identifying whether provided demonstrations are consistent with previously provided demonstrations. While this could help a learner avoid taking demonstrations from a bad teacher, this does not directly evaluate the quality of teaching and so does not necessarily help the teacher improve.

The proposed framework therefore intends to emphasise the teacher's contribution to learning, and provide new approaches for designing feedback and evaluation tools in \LfD systems.

\subsection{Insights from Machine Teaching}\label{s:machine_teaching}
The ability to analyse teaching behaviour is critical to improving \LfD performance. It can be shown that an optimal teacher can theoretically provide the minimum number of samples required to teach a learner a task, called the \textit{teaching dimension} \cite{Goldman1995OnTeaching,Balbach2009RecentTeaching,Khan2011HowDimension,Cakmak2014,Zhu2015MachineEducation}, by providing \textit{non-i.i.d.} samples to the learner which exploit the task structure and learning method employed. In toy-problems, the learning method and task structure can be clearly defined for analysis, but when using \LfD for learning real-world tasks from novice users this is rarely the case and so the analysis methods employed in machine teaching are not directly applicable.

While identifying theoretically optimal teachers may not be possible in general \LfD tasks, it is shown that providing feedback and guidance to the teacher to help them become informed about the learning process can change participants' teaching strategy, resulting in improved teaching performance \cite{Zang2010BatchDemonstration, Cakmak2014}. It is this feedback mechanism for improving the teacher that the here proposed framework aims to enable, by providing a structure with which \LfD problems can be analysed, and with which support tools for teaching can be designed.

\subsection{Modelling Learning from Demonstration}
There have been a number of efforts in the wider \newabbreviation{\HRI}{HRI} domain to model human behaviour and cognition when interacting with robots. These have typically focused on collaborative human-robot teams that work together to complete a task, rather than the problem faced in \LfD of teaching new skills \cite{Goodrich2007,Nikolaidis2017Human-RobotAutonomy, Hiatt2017HumanCollaboration}.

Looking more specifically at \HRI involving uncertain, or noisy, communication, \citet{Hellstrom2018UnderstandableHow} present a more general framework based on message compression. Here they discuss the robot's ability to influence the person through communication, however, being a general framework for \HRI, it serves as a complementary resource to the more focused \LfD-specific framework presented here. 

While there have been several works highlighting the benefits of considering the teacher as an active contributor to the learning process in \LfD, as discussed, there is a lack of structure in considering their role. A significant previous effort to introduce a formal structure to \LfD can be found in \cite{Billing2010}, where the authors model the \LfD process following a message compression scheme. While this helps to provide a framework around the learning process in \LfD, there is no consideration for how interaction with the system itself affects the teacher's actions. By not considering the influence of the teacher, the current approaches of focusing on the learner neglects opportunities for improving task learning performance in \LfD. 

An extension from \citet{Billing2010} is considered in \citet{Cederborg2014ADo}, where the authors present a generalised framework which considers multiple modalities of teacher feedback to the robot, with the view of using ambiguous teaching cues such as gaze to infer the learner's goal. Though the teacher is assumed to be infallible, the learner is presented as a passive receiver of information from the teacher in this case, missing the opportunity to actively improve the teacher's understanding of the learning process beyond explicit training by an expert.

Given the lack of a distinct framework which captures the teacher's contribution to the \LfD process explicitly, and the benefit of doing so as suggested in prior works, there is a clear need to develop an improved formalism. By considering the teacher explicitly, it would be possible to define teaching objectives and quality measures in a task-agnostic representation, and design tools that allow the robot to more effectively guide the teacher toward better demonstrations. 

\section{Learning and Teaching Framework}\label{s:theory}
In this section, a new, formal framework for assessing the role of the teacher in robot \LfD is defined, along with the means by which it can be used to measure, evaluate, and modify teacher behaviour. The proposed framework builds on the prior work of  \citet{Billing2010}, but diverges in that it places much greater emphasis on the role of the teacher.

\begin{figure*}[ht!]
\centering
	\def\svgwidth{0.95\linewidth}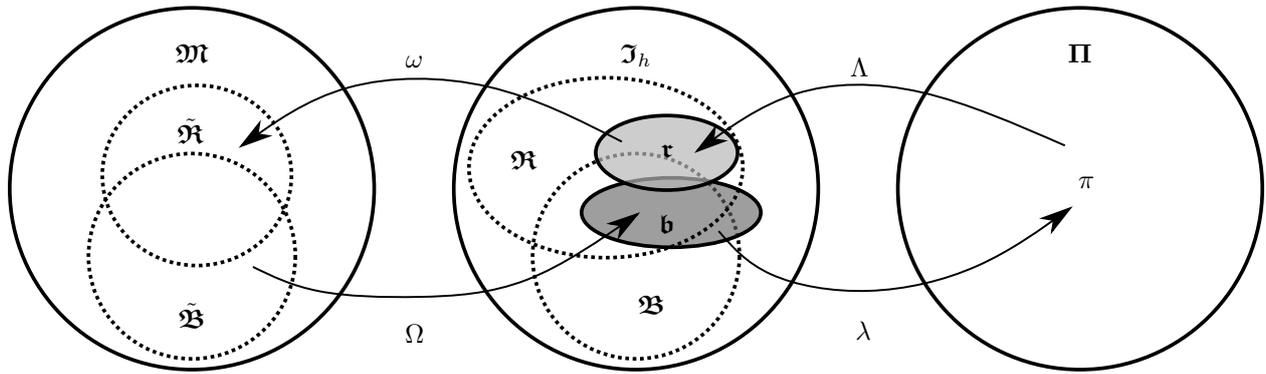
	\caption{Learning from Demonstration model incorporating imperfect teaching. The space introduced on the left, $\spaceM$, represents the user's understanding of what the robot has learned so far, and what they would like the robot to learn. The shaded regions indicate parts of the process which are directly observable, \ie the actual demonstrations provided $\spaceb$ and the task realisations $\spacer$.}
	\label{f:lfd_model}%
\end{figure*}

\subsection{Framework Definition}\label{s:framework_def}
The proposed framework describes the \LfD process as a set of mappings between an information history space, $\spaceIh$, a policy space, $\spacePi$ and a teacher belief space $\spaceM$. $\spaceIh$ represents the space of all possible \textit{event histories}, where each element is a tuple containing initial conditions, observations, and action sequences describing a full rollout of some behaviour (\ie task), . $\spacePi$ represents the space of \textit{all possible policies, learnable by a selected learning function} $\lambda$. $\spaceM$ captures the \textit{teacher's belief} about the extent to which the robot has learnt the task, and the demonstrations necessary to improve its performance. An overview of the framework is presented in \fref{lfd_model}, and the following highlights some of its salient features.

\subsubsection{Task Teaching ---}
\renewcommand  {\Nk}     {M} 
\renewcommand  {\nk}     {m} 
\providecommand{\cSpaceb}{|\spaceb|} 
The space representing a behaviour, or task, that the user would like to teach the robot is denoted $\spaceB$, the \textit{task space} (see \fref{lfd_model}). 

The elements of $\spaceB$ represent all possible ways the robot should be capable of performing the task. For instance, in the plant grading task in \fref{gex}, $\spaceB$ might represent the set of all possible trajectories for picking and disposing of plants from any location in a tray. The learner does not have access directly to $\spaceB$, instead relying on a set of $\Nk$ demonstrations provided by the teacher, $\spaceb$. In the plant grading task example, $\spaceb$ consists of the set of demonstrated trajectories for picking plants from specific locations (cyan trajectories in \fref{gex}). The goal of improving teaching efficiency is therefore conceptually captured in this framework as that of \emph{providing the best possible $\spaceb$ to enable the robot to learn the target behaviour $\spaceB$}. A formal metric for this is discussed below (refer to \sref{evaluating_teaching}).

\subsubsection{Task Learning ---}\label{s:task_learning}
The framework assumes that the robot is equipped with learning capability, such that it can use demonstrations to form a task model. Specifically, there exists a learning function, $\lambda$ that uses $\spaceb$ to derive a controller $\policy=\lambda({\spaceb})\in\spacePi$.

In real-world tasks, it is typically the case that\footnote{Throughout this paper, $|\boldsymbol{\mathfrak{a}}|$ is used to denote the cardinality of the set $\boldsymbol{\mathfrak{a}}$.} $\cSpaceb\ll|\spaceB|$, therefore the learning system must learn a model which can \textit{generalise} from the subset $\spaceb$ provided. It is also typical that learning occurs as an \emph{iterative operation}, that is, the learnt model is sequentially updated whenever new demonstrations are provided. The framework captures this by assuming that $\lambda$ makes use of all past demonstrations provided, \ie $\policy=\lambda({\spaceb_\nk})\in\spacePi$, where $\spaceb_{\nk}$ contains the $\nk$ demonstrations made available.

\subsubsection{Task Realisation --- }\label{s:task_execution}
Once a controller has been learnt, it can be executed by the learner to (try to) perform the task. Execution of a task using a learnt model is denoted as a mapping from $\spacePi$ to $\spaceIh$ through a ``realisation function'' $\Lambda$, where each execution results in a \textit{task realisation}, $\spacer=\Lambda(\policy)\in \spaceIh$. Here, $\spacer$ represents the set of task realisations which are \textit{actually observed} by the teacher. In the plant grading example task, these realisations, would represent the robot using its learnt model to generate picking trajectories (shown in red in \fref{gex}). The realisation space, $\spaceR$, represents the set of all possible realisations of the learnt model under different task conditions. It is desired that the final learnt model would result in $\spaceB\subseteq\spaceR$, which is the case that the target task has been learnt completely.\footnote{Strict equality is not required, as $\spaceR \setminus \spaceB$ represents event histories which do not occur during the target task.} Note, however, it is typically the case that $|\spacer|\ll|\spaceR|$. This means that $\spaceR$ is not readily observable in practice, hence measurement of learning performance may require approximations.

\subsubsection{Generalisation ---}\label{s:generalisation}
In order for the learner to learn the target task effectively and efficiently, it is required for the learner to \textit{generalise} from the provided demonstrations. Under the presented framework, generalisation is defined as 
\begin{equation}
(\spaceR\setminus\spaceb)\cap\spaceB\neq\emptyset.
\label{e:generalisation_definition}
\end{equation}
In other words, there exist task reproductions that do not occur in the demonstrated data $\spaceb$, but do fall within the definition of successful task performance $\spaceB$. This is captured by the striped region in \fref{model_failures}.

\subsubsection{Modelling the Teacher --- }\label{s:modelling_the_teacher}
\providecommand{\espaceR}{\estimated{\spaceR}}
\providecommand{\espaceB}{\estimated{\spaceB}}
To develop an intuition for teaching behaviour and, importantly, why the teacher can fail to provide adequate demonstrations, the framework introduces a third space modelling the \emph{teacher's beliefs about the learning process}. This is represented by a \textit{belief space}, $\spaceM$ (see \fref{lfd_model}). 

As the teacher has no direct way of knowing the learner's state, $\spaceR$, they must estimate it based on the available information, \ie the realisations that they observe, $\spacer$. This information is combined with the teacher's prior knowledge and biases, $\Hfact$, to form an \textit{interpretation} of the learner state $\espaceR$. This is modelled as a mapping from $\spaceIh$ to an estimated realisation space $\espaceR \subset \spaceM$ through an interpretation function $\espaceR=\omega(\spacer,\Hfact)$. 

In the context of iterative teaching, this estimated realisation space is then used to guide the teacher's next demonstration, in conjunction with their internal idea of what task the robot must learn $\espaceB$, and their internal biases $\Hfact$. This is modelled as a mapping from $\spaceM$ back into $\spaceIh$ through a demonstration function
\begin{equation}
\spaceb_{\nk+1} = \Omega(\espaceR,\espaceB,\spaceb_\nk,\Hfact).
\label{e:demo_fn}
\end{equation}
With the addition of $\spaceM$, it is possible to close the loop on the \LfD process with a consideration for the teacher's internal belief processes and begin to reason about how teaching failures occur in this iterative cycle, see \fref{pipeline}.

\providecommand{\aspaceR}{\hat{\spaceR}}
\providecommand{\aspaceB}{\hat{\spaceB}}
\subsubsection{Objective Task Approximations}\label{s:approx}
If the task space is very large or continuous, some form of approximation $\aspaceB\sim\spaceB$ and $\aspaceR\sim\spaceR$ may be required for measurement of teacher performance in practical situations.
Note that the approximations $\aspaceB$ and $\aspaceR$ are distinct from the teacher's internal evaluation of the task and learner performance ($\espaceB$ and $\espaceR$, respectively) since they are \emph{objective}, and therefore independent of the individual teacher's biases (\ie are not influenced by $\Hfact$).

Such approximations may be formed through several approaches, from state-space reduction via discretisation and bounding, to statistical approximation with Monte Carlo methods. For example, in the plant grading task (shown in \fref{gex}), $\aspaceR$ can be defined as the trajectories generated given a pre-defined set of `test plant' locations in the tray, enabling a measure of the learnt policy's performance to be made, without having to exhaustively test every possible plant location and every admissible approach trajectory for any given plant.

The following sections describe measures of teaching performance and teaching failure modes in the general case of $\spaceB$ and $\spaceR$, however, these are equally applicable in the case that approximated spaces are used. 

\begin{figure}[t]
	\centering
	\def\svgwidth{\linewidth}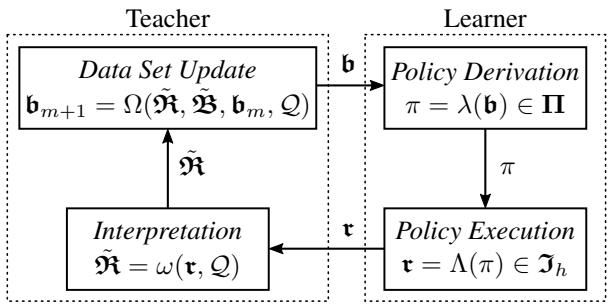
	\caption{\LfD pipeline, accounting for teacher influence. The teacher's interpretation of the robot's performance, $\espaceR$ is based on the task realisations that they actually observe, $\spacer$, and hidden human factors (internal biases), $\Hfact$. The way in which the teacher provides new demonstrations then depends on this interpretation, their interpretation of target behaviour $\espaceB$, and $\Hfact$.}
	\label{f:pipeline}%
\end{figure}

\subsection{Evaluating Teaching by Demonstration}\label{s:evaluating_teaching}
While evaluating the performance of the \emph{learner} in \LfD is well established, relatively little attention has been given to the performance of the teacher. The primary objective of the teacher is to ensure the learnt model allows the robot to execute the target skill.  This primary objective is formally defined using the proposed feedback, along with further measures which are useful for evaluating teacher performance and identifying teaching failures. 

\subsubsection{Teaching Efficacy --- }\label{s:efficacy}
The primary objective of teaching is to have the robot accurately learn the target skill, such that it is able to perform the desired task. From the perspective of \emph{learning}, a necessary condition for this is that $\spacePi$ contains a model sufficiently close to the target skill, and that $\lambda$ applied to the demonstrations $\spaceb_\nk$ correctly identifies this model.

\citet{Billing2010} posit that the \emph{learning objective} can be described as the minimisation of $\spaceB\setminus\spaceR$ and $\spaceR\setminus\spaceB$. In other words $\spaceB$ and $\spaceR$ should overlap exactly, meaning that the robot would reproduce the skill completely, and exclusively to, the entire space of the target behaviour. However, in practice, this objective is both \il{\item\label{i:helstrom-problem-i} difficult to achieve (since in many cases, the spaces $\spaceR$ and $\spaceB$ are too large to fully observe, refer to \sref{task_execution}), and \item\label{i:helstrom-problem-ii} overly prescriptive of the robot's behaviour.} Specifically, in terms of the latter, it is often the case that learner behaviour in conditions outside the desired task, \ie $\spaceR\setminus\spaceB$, can simply be ignored. For example, in the plant grading task in \fref{gex}, the robot might learn how to pick plants \textit{in close proximity, but external, to the tray} as a consequence of learning to pick those in the tray edge positions (a case of $\spaceR\setminus\spaceB\neq\emptyset$). This, however, does not conflict with performing the target skill, nor increase teaching effort. 

Hence, the metric proposed here to evaluate \emph{teaching efficacy} is
\begin{equation}
\efficacy = \frac{|\spaceR\cap\spaceB|}{|\spaceB|},\quad \efficacy \in [0,1].
\label{e:efficacy}
\end{equation}
In other words, the goal is to achieve $\spaceB\subseteq\spaceR$, while ignoring what the model has learned outside of the target space (\ie $\spaceR\setminus\spaceB$), normalised by the size of the task space $\spaceB$. This measure can be treated as an objective measure of the teacher's ability in enabling the learner to acquire the skill needed for the task.

The teacher's performance can then be monitored during interaction with the robot by considering the difference in efficacy with each additional demonstration.

\subsubsection{Teaching Efficiency --- }\label{s:efficiency}
Having defined the measure of teaching efficacy, it is then possible to consider the \emph{teaching efficiency}. Efficiency in any given application is often context-dependent, however, typically, it is desirable to \emph{minimise the total number of demonstrations required} since this is correlated with both the time spent teaching and the space needed to store data.

With this in mind, \emph{teaching efficiency} is defined here simply as efficacy normalised by the number of demonstrations provided
\begin{equation}
\efficiency = \frac{\efficacy}{|\spaceb|}, \quad \efficiency \in [0,1].
\label{e:teach_eff}
\end{equation}
In other words, to be efficient, the teacher must achieve the maximum efficacy with the fewest possible demonstrations. Importantly, \emph{ambiguous demonstrations}, \emph{undemonstrated states}, and \emph{incorrect demonstrations} will all result in reducing the teaching efficiency metric, as discussed below.

\vspace{2ex}
\noindent The definitions \eref{efficacy} and \eref{teach_eff} provide a way to monitor a user's teaching performance. In the next section, commonly-encountered teaching failures are analysed in light of the new framework.


\subsection{Understanding Teaching Failures}\label{s:teaching_failures}
With the framework established, it is now possible to make formal definitions of failure modes in \LfD teaching, and metrics for their quantitative evaluation. Specifically, the below examines three common teaching failures, namely, \il{\item \emph{incorrect demonstrations}, \item \emph{ambiguous demonstrations}, and \item \emph{undemonstrated states}}. Each of these can be attributed to poor teacher skill, $\Omega$, affecting the quality of data provided to the learner, or poor user judgement, $\omega$, affecting the accuracy of the teacher's estimation $\espaceR$ (see \sref{modelling_the_teacher}).

\begin{figure}[t]%
	\centering%
	\def\svgwidth{\linewidth}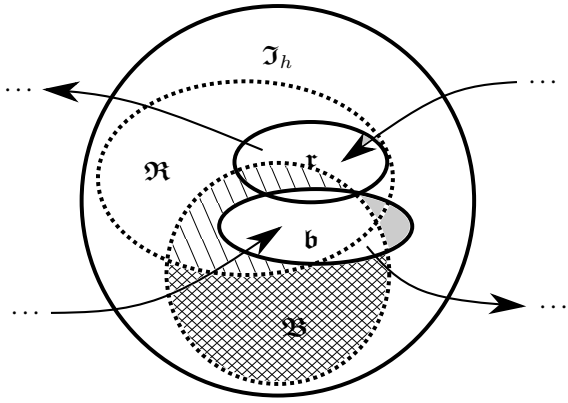%
	\caption{Focusing on the information history space, three subsets of interest can be identified. The striped region in the centre of the diagram represents parts of the task which have been correctly learned through \textit{generalisation} from the demonstration set. The cross-hatched region then represents undemonstrated regions of the task. Finally the shaded region represents demonstrations that the teacher has performed incorrectly. See \sref{evaluating_teaching}.}
	\label{f:model_failures}%
\end{figure}

\subsubsection{Incorrect Demonstrations ---} \label{s:wrong_demos}
The most fundamental teaching failure is that of providing \emph{incorrect}, or unsuccessful, demonstrations, \ie demonstrations that are \emph{not} examples of the target skill. Formally, these can be defined as
\begin{equation}
\spaceb\setminus\spaceB\neq\emptyset
\label{e:wrong_demo}
\end{equation}
illustrated by the solid-shaded region in \fref{model_failures}.

The effect of incorrect demonstrations can be observed through a reduction in teaching efficacy, as the learner's performance will degrade
\begin{equation}
\efficacy_m - \efficacy_{m-1} \leq 0.
\label{e:incorrect_demo}
\end{equation}

\noindent%
Incorrect demonstrations can occur for a number of reasons. For instance, the teacher may have poor skill in operating the robot, possibly indicating they need more time interacting with the robot to become accustomed to it. Alternatively, the teacher may make errors, such as forgetting which object they had decided to demonstrate for the robot, mid-demonstration. In both cases, incorrect demonstrations represent a failure in the demonstration function, $\Omega$, as a result of human factors specific to the teacher, $\Hfact$, in the provision of demonstration data (see \fref{pipeline}). While it may be difficult to pinpoint a particular reason for an incorrect demonstration, \eref{incorrect_demo} provides a measure for identifying an incorrect demonstration.

\subsubsection{Undemonstrated States ---} \label{s:failures_undemo}
As mentioned in \sref{task_learning}, usually $|\spaceb|\ll|\spaceB|$, so it is inevitable that the learner must generalise as much as possible from the demonstration set to effectively perform the desired task. If there are relevant states in which the robot cannot perform the task, after the teacher has provided all of the demonstrations they \emph{think} are required for the robot to learn, these remaining states are referred to as undemonstrated states.

Undemonstrated states occur as there is a limit to the extent which generalisation is possible for a given $\spaceb$, and so the teacher must form good estimations of the robot's learned ability to identify when enough data has been provided, \ie ideally $\espaceR\sim\spaceR\sim\spaceB$.

The formalism defines undemonstrated states as the set
\begin{equation}
\spaceB\setminus(\spaceR\cup\spaceb)\neq\emptyset
\label{e:undemonstrated-e2}
\end{equation}
illustrated by the cross-hatched region in \fref{model_failures}.

From the perspective of the teacher, failure to adequately demonstrate the task, such that the learner is left with undemonstrated states that prevent it performing the task is a sign that the teacher does not understand the learner's current abilities, \ie they have a poor estimate of $\espaceR$ where they are \emph{overestimating} the learner's ability to perform the task. Pathways to improving this estimate may include improving the selection of task realisations $\spacer$ to give more useful information to the teacher, or explicitly training the teacher on how the robot learns in order to give them more accurate expectations, $\Hfact$, of the learner.

\subsubsection{Ambiguous Demonstrations ---}\label{s:failures_ambig}
The third teaching failure that commonly occurs in \LfD scenarios is the issue of \emph{ambiguous demonstrations}. These are defined as demonstrations which offer little or no new information to the learner, such that performance on the target task is not improved. This can happen, for example, when demonstrations provided by the teacher are very similar to those already been seen by the learner. 

With this in mind, an ambiguous demonstration is defined as one which is \emph{similar} to demonstrations already provided, within some threshold
\begin{equation}
s(b,\spaceb) \leq \delta_a,
\label{e:ambig_sim}
\end{equation}
where $b$ is the new demonstration, $s(\cdot,\cdot)$ is some measure of similarity of $b$ with respect to the existing demonstrations in $\spaceb$, and $\delta_a$ is an ambiguity threshold level for $s$. For instance, $s$ could be the mean Euclidean distance difference between a demonstrated trajectory and each of those in the data set $\spaceb$, and $\delta_a$ set at some minimum threshold for this.

Note that, provision of ambiguous demonstrations results in little or no improvement in the learner's performance, but should also not significantly degrade it. Therefore, the effect of ambiguous teaching is that the change (or lack thereof) in learner efficacy lies within an upper and lower bound
\begin{equation}
\delta_l \leq \efficacy_m - \efficacy_{m-1} \leq \delta_u.
\label{e:ambig_alt}
\end{equation}

\noindent%
Similar to undemonstrated states, ambiguous demonstrations can be attributed to the teacher forming a poor estimate of the learner's ability, $\espaceR$, however, in this case, the teacher is \emph{underestimating} the impact individual demonstrations are having on the learner's ability.

\vspace{2ex}\noindent%
To summarise, the proposed framework improves upon the previous models of \LfD by explicitly modelling what is observed by the human teacher, and how this might be interpreted by them. With this modification, definitions for measures of teacher performance and teaching failures naturally follow. With these performance measures in place, it is possible to design feedback tools with the specific purpose of influencing the teacher's belief space, $\spaceM$, which can be used to guide teachers toward more effective teaching practices. This is explored in the following two experiments, demonstrating the benefits of the proposed framework.

\section{Evaluation}\label{s:evaluation}
In this section, two experimental studies in \LfD are presented to explore how the proposed framework can be used to measure, evaluate, and modify teacher behaviour to improve the overall system performance. Specifically, the experiments examine how it allows the design of effective feedback approaches that help the teacher understand the learner by improving interpretation, $\omega$, to give the teacher better estimations of the learner's ability, $\espaceR$. By closing this loop between teacher and learner, it is hoped that the teacher will be able to provide higher quality demonstrations, $\spaceb$, during the data set update step, and thus increase the performance of the robot compared to an unguided teacher.

\subsection{Experiment 1: Point-to-Point Reaching}\label{s:expt-ptp}
In the first experiment, the teacher must teach a robot to navigate a maze from a start region to a goal (see \fref{frames:a}). This task is chosen as it represents a relatively simple challenge in terms of robot \emph{learning}, but the performance of the learner depends critically on the teacher's skill in understanding what the robot has learned from previous demonstrations. 

In particular, the efficiency and efficacy of teaching depends on the teacher's ability to ensure the robot \emph{generalises from the teaching examples given}, such that it is able to generate a path from anywhere within the designated start zone to the goal. In other words, $\spaceB$ represents the space of all such admissible trajectories , and the robot must learn to approximate this from the subset $\spaceb$ provided by the teacher. To do this, the teacher must form a belief $\espaceR$ of the robot's actual ability $\spaceR$ before selecting the demonstration set $\spaceb$.
 
It is expected that the teaching quality can be modified and improved by manipulating the teacher's belief space $\spaceM$, specifically, their prior beliefs about the learner, $\Hfact$, and their estimate of the robot's actual ability $\espaceR$. Due to the relatively low-dimensional nature of the problem, it is hypothesised that this can be achieved through appropriate design of visual feedback. The following describes the experimental procedure for evaluating this hypothesis and reports results from a group of novice teachers\footnote{This experiment was approved by a KCL ethics committee, ref. LRS-16/17-3800. Informed consent was obtained from all experimental participants. The data collected for this research is open access, with accreditation, from {\tt\small http://doi.org/10.18742/RDM01-242}.}.

\begin{figure}%
	\begin{subfigure}[t]{0.58\linewidth}
		\captionsetup{justification=centering}
		\includegraphics[height=3.5cm,trim={0cm 1.2cm 12cm 0cm},clip]{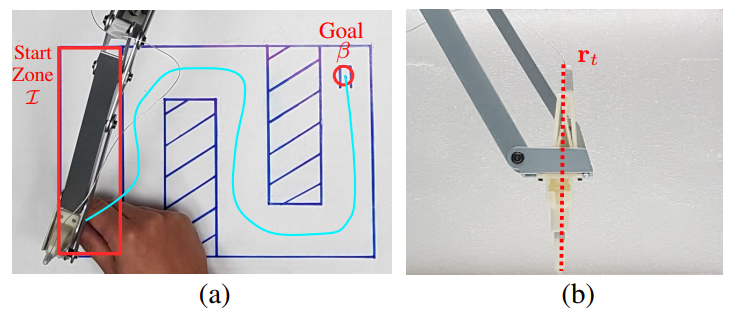}
		\caption{~}
		\label{f:frames:a}
	\end{subfigure}
	\begin{subfigure}[t]{0.25\linewidth}
	\captionsetup{justification=centering, margin=0.7\linewidth}
		\includegraphics[height=3.50cm,trim={15cm 1.2cm 2cm 0cm},clip]{setup.png}
		\caption{~}
		\label{f:frames:b}
	\end{subfigure}
	\vspace{-0.4cm}
	\caption{Experimental set up. 
		\begin{enumerate*}[label=(\alph*)]
			\item Participants are asked to guide a lightweight robot through the work space from the start zone to the goal location, avoiding the maze boundaries and the two obstacles (shaded zones).
			\item End-effector of the robot indicating the positioning of the TrakStar sensor used for recording data.
		\end{enumerate*}
	}
	\label{f:frames}%
\end{figure}

\begin{figure*}
	\def\svgwidth{\linewidth}
	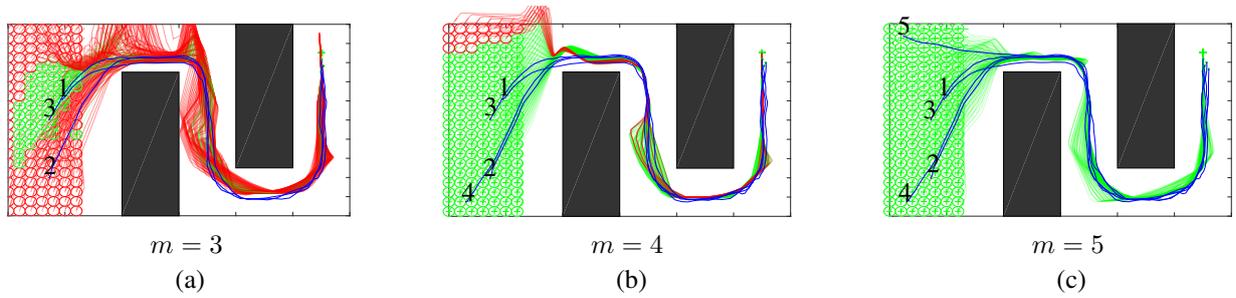
			\put(-411,-13){(a)}
			\put(-246,-13){(b)}
			\put(-81,-13){(c)}
			\vspace{-.2cm}
		\caption{Example of a demonstration sequence for $3\le \nk\le5$ demonstrations, and the visualisation shown to participants during the visual feedback test conditions. Trajectories in $\aspaceR$ are illustrated, with those that meet the criteria \eref{exp1criteria} coloured green and those that do not coloured red. The teacher's demonstrations (numbered in the order in which they are given) are overlaid in blue.}%
	\label{f:demos}%
\end{figure*}

\subsubsection{Hypotheses ---}\label{s:hypotheses-1}
 The experimental hypotheses are chosen to test whether teacher skill, as measured by the metrics \eref{efficacy} and \eref{teach_eff}, can be improved through appropriate  feedback to the teacher. They are formally defined as:
 \begin{hypotheses}
 	\item \label{E1H1} \textit{Visualisation of the robot's learning progress results in a significant improvement in teaching efficiency, compared to a no-guidance teaching process.} 
 	\item\label{E1H2} \textit{A heuristically guided teaching process, which uses visual feedback plus a rule set, results in significantly improved teaching efficiency, compared to a no-guidance teaching process.}
 	\item\label{E1H3} \textit{A heuristically guided teaching process, which uses visual feedback plus a rule set, results in significantly improved teaching efficiency, versus a solely visually guided teaching process.}
 \end{hypotheses}
 ~
 \\
\ref{E1H1} tests whether knowledge of the learner's actual ability $\aspaceR\sim\spaceR$, improves teacher/learner performance beyond the performance achieved when relying on the teacher's understanding $\espaceR$ alone. 

\ref{E1H2} introduces a heuristic rule set for user guidance, testing whether influencing the teacher's biases, $\Hfact$, along with providing knowledge of $\aspaceR$ improves teacher/learner performance beyond relying on $\espaceR$ alone. 

\ref{E1H3} considers whether modifying $\Hfact$ along with providing $\aspaceR$ improves teacher/learner performance beyond just the feedback of $\aspaceR$ alone. 

\subsubsection{Materials and Methods ---}\label{s:setup-1}
Participants in the experiment are asked to teach a robot to navigate its end-effector through a simple two-dimensional maze. To do this, they must provide demonstrations to the robot by gripping its end-effector and manually guiding it through the maze, while avoiding the workspace boundaries and obstacles (see \fref{frames:a}).

More formally, the task skill to be taught to the robot is defined as generating a path which, beginning in a starting area $\Is$, passes through an \textit{admissible space} $\spaceX$ to reach a target $\Ts$, see \fref{frames:a}. $\spaceX$ is defined by a two dimensional bounding rectangle ($20\,cm$ by $30\,cm$), containing the start zone, $\Is$ (a $20\,cm$ by $6\,cm$ rectangle) and the target, $\Ts$ (a $0.5cm$ diameter circle). The two obstacles (shown in the figure as shaded blocks) are \textit{not} included in $\spaceX$. Any trajectory $\traj_{\nk}$ that links points in $\Is$ to $\Ts$, without leaving $\spaceX$, is a member of $\spaceB$ 
\begin{equation}
\quad \traj \in \spaceB \quad \textrm{if} \quad
\begin{cases}
\textnormal{(i)} \quad &\traj \subset \spaceX \\
\textnormal{(ii)} \quad &\traj(0) \subset \Is \\
\textnormal{(iii)} \quad &\traj(T) \subset \Ts
\end{cases}
\label{e:exp1criteria}
\end{equation}
where $\traj_m(0)$ and $\traj(T)$ represent the first and last sample in the recorded trajectory respectively. Note that these criteria apply to both demonstrations and task realisations.

The robot learner used in this experiment is a uFactory uArm Metal, a lightweight robot ($<\!1\,kg$) with back-driveable motors. During teaching, the robot end effector position is recorded using an NDI TrakStar sensor, (see \fref{frames:b}). This provides $\pm1.3\,mm$ RMSE positioning accuracy at a sampling rate of $80Hz$.

Using this setup, a data set consisting of $\Nk$ demonstrations is collected during teaching. Each demonstration consists of a trajectory containing $T_\nk$ samples of the robot state, $\statept_\nd$, (giving a total of $J=T_\nk \times \Nk$ samples).

Using this data, the robot uses a \newabbreviation{\TPGMM}{TP-GMM}  \cite{Calinon2015} to learn a model of the demonstrated task. While there are many learning methods which are suitable for this step, \TPGMM is chosen as it has been shown to be particularly effective in generalising from a limited set of demonstrations to unseen conditions. It is important to note that with \TPGMM the robot learner cannot self-refine the learned model, \ie the model is only as good as the data provided by the teacher. Therefore the teacher must make good assessments of the learner's ability, $\espaceR$, in order to provide suitable set of demonstrations. 

In \TPGMM, the task is parameterised by a collection of affine transformations which, in this case, represent a collection of reference frames marking robot end-effector locations and object locations. A local mixture model is learned for the demonstration data in each of these frames of reference. The local models are then combined to the global frame of reference through a product of Gaussians, resulting in a trade-off between the local models which optimises the consistencies observed in data in each frame of reference. Continuous trajectories can then be generated from the global model using Gaussian Mixture Regression (the reader is referred to \cite{Calinon2015} for full details of the \TPGMM). In the implementation used here, the model contains $K=11$ Gaussian components, and the state  $\statept_\nd=(t_\nd,\mathbf{x}_\nd^\T)\in\mathbb{R}^{3\times1}$ is represented with the time $t$ and end-effector position $\mathbf{x}$ for sample \nd. Four sets of task parameters  are used, defining the start, end, and obstacle locations. 

Teacher performance is measured according to the efficacy \eref{efficacy} and efficiency \eref{teach_eff} of teaching. Here, as $\spaceB$ and $\spaceR$ represent continuous spaces in this task, it is necessary to define objective approximations for quantitative comparisons (see \sref{approx}). In this experiment, the goal region $\Ts$ is small, so the primary task of teaching is to ensure the learner achieves good generalisation over $\Is$. A simple way to approximate $\spaceB$, therefore, is to discretise $\Is$ into a finite set of points, and consider any trajectory that links one of these points to the target $\Ts$, while meeting the criteria \eref{exp1criteria}, as an element of $\aspaceB$. In the results reported here, $\Is$ is discretised into a grid of $20\times7$ points, so $|\aspaceB|=140$. Similarly, $\spaceR$ is approximated by the set of trajectories generated by the \TPGMM from each of these points, therefore maximum performance is achieved when $|\aspaceR\cap\aspaceB|=140$ giving $\efficacy=1$, as shown in \fref{demos}(c).

In two of the experimental conditions, visual feedback is provided to experimental subjects (refer to \sref{hypotheses-1} and \sref{procedure-1}). As a simple means to generate such feedback, the trajectories in $\aspaceR$ are presented to the subjects, overlaid onto an image of the maze. These are coloured green if they meet the criteria \eref{exp1criteria} and red otherwise, as shown in the examples in \fref{demos}.

\subsubsection{Procedure ---}\label{s:procedure-1}
The following describes the protocol for working with experimental participants. 

The experiment is designed as a within-subjects study, so a repeated-measures Analysis of Variance (ANOVA) study is used to determine if any significant effects can be observed in the data. The independent variable in each phase of the experiment is the level of guidance provided to the participant. In all stages of the experiment, the dependent variable is the participant's teaching efficiency, $\eta$.

A power analysis for a repeated measures ANOVA, with one group and three measurement levels, indicates that, for a medium effect size (Cohen's $f=0.3$) and a power of $0.95$, the required sample size is $\Ns=30$ \cite{Faul2009StatisticalAnalyses}. In the results reported below, therefore, are for the experiment conducted with $30$ participants ($17$ male, $13$ female; ages $\mu=35.1$, $\sigma=9.7$). All participants are pre-screened to ensure that they have no background in robotics or machine learning.

The experiment commences with the subject watching an introductory video that \il{\item explains \LfD using a real-world task example, \item introduces the objective of teaching the robot to generate trajectories from the start zone to the goal through the maze}. Videos are used for all instruction to ensure consistency in the participants' prior knowledge and understanding of teaching phase\footnote{The introductory video used in the experiment is provided as a supplementary file to this paper, and is available to view online \mbox{\url{https://youtu.be/eafZ1bRAGLM}}}. After the introductory video, they are given one minute to familiarise themselves with the robot, where they are free to move it around. After this, the main experiment begins. This is split into three phases, one for each test condition. Each phase is introduced by a video providing instructions, and the participant is given one attempt to provide a set of demonstrations.

\subsubsection{Conditions --- } The following are specific details on each test condition used during the experiment. 

\textit{Condition \ref{E1C1} - No Feedback (NF):}\manuallabel{E1C1}{1}
In this phase, participants are tested to see if they are able to provide demonstrations with no feedback, \ie $\spacer=\emptyset$, thus relying only on their (uninformed) expectations of the system's behaviour, $\Hfact$. The instruction video explains the different areas of the task map, provides one basic example of how the task is meant to be performed, and explains that they must provide as many demonstrations as they feel are necessary to enable the robot to perform the task from any point in the start zone.

\textit{Condition \ref{E1C2} - Visual Feedback (VF):}\manuallabel{E1C2}{2}
In this phase, the effect of providing a transparent visualisation of the robot's learning progress is tested. In between each demonstration, the user is provided with the visualisation of learning progress, described in \sref{setup-1} and shown in \fref{demos}. The instruction video explains the visualisation, and provides a simple example of what demonstrations look like in the visualisation.

\textit{Condition \ref{E1C3} - Visual Feedback, Rule Guidance (VR):}\manuallabel{E1C3}{3}
In this phase, the participant must additionally follow a set of rules when providing their demonstrations. The rule set is designed to approximately guide users to avoid undemonstrated states and ambiguous demonstrations. The rules are \il{\item provide one demonstration, starting from anywhere in the start zone, 
\item provide demonstrations within $4\,cm$ of the starting point of the first demonstration, until it is surrounded by successful (\ie green) test trajectories, and
\item continue providing demonstrations within $4\,cm$ of the starting point of successful trajectories, in the area with the greatest number of failed (\ie red) trajectories.} The instruction video explains the rules through a single simple example, but avoids dictating exactly how the participant should teach by only showing a small section of the start zone when explaining the rule set.

\begin{figure}[t]
	\def\svgwidth{\linewidth}\input{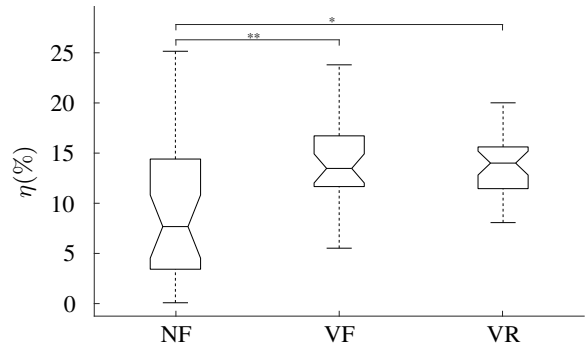}      
	\caption{Box plot of results for each test condition. No Feedback (\EICI), Visual Feedback(\EICII), Visual Feedback/Rule Guidance (\EICIII). The red lines indicate median values of the teaching efficiency, $\eta$, in each case. The bottom and top of the blue boxes indicate the $25^{th}$ and $75^{th}$ percentiles, respectively. The upper and lower of the dashed lines indicate maximum and minimum values, respectively. }
	\label{f:icra_results}      
\end{figure}
\begin{figure}[t]
	\def\svgwidth{0.95\linewidth}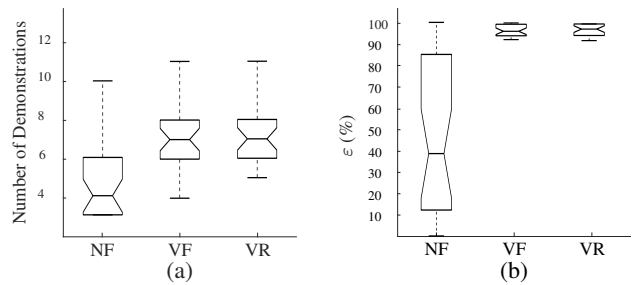
	\manuallabel{f:icra_results2:a}{(a)}%
	\manuallabel{f:icra_results2:b}{(b)}%
	\caption{{Breakdown of} \ref{f:icra_results2:a} the number of demonstrations provided in each phase, and \ref{f:icra_results2:b} the percentage of successful trajectories generated, \ie the \textit{efficacy}, $\efficacy$.}
	\label{f:icra_results2}      
\end{figure}

\subsubsection{Results ---}\label{s:results-1}
Prior to performing a repeated measures ANOVA analysis, the model assumptions were all checked and verified for the collected data; continuous dependent variable, at least two groups of independent variables, absenece of outliers, normally distributed dependent variable, and sphericity of data.

The number of demonstrations used for evaluating user performance was taken to be the number of demonstrations required to achieve at least 90\% coverage of the test grid, or the maximum number of demonstrations if 90\% coverage was not achieved. Analysing the teaching efficiency score, the data indicates a significant effect of the feedback method on the teaching efficiency, $F(2,58)=7.952, p=0.001$. As a significant effect was observed, a multiple comparisons of means was performed. A significant difference was found between the \EICI and \EICII conditions, $p = 0.006$, as well as between the \EICI and \EICIII conditions, $p = 0.017$. No significant difference was observed between the \EICII and \EICIII conditions, $p = 0.801$. The median teaching efficiencies, $\eta$, are shown in \fref{icra_results}. The difference in teaching efficiency medians between the \EICII ($13.4\%$) condition and the \EICI ($7.6\%$) condition shows a relative improvement of approximately $180\%$. A similar teaching efficiency improvement can be observed between the \EICIII ($14\%$) condition and \EICI.

These results and the statistical analysis show support for \ref{E1H1} and \ref{E1H2}. This shows a clear benefit in providing the visual feedback to the teacher during \LfD, \ie by helping the teacher to gain an accurate understanding of the learner's current ability, $\espaceR$, teaching is improved. There is no support found for \ref{E1H3}. As seen in \fref{icra_results}, the teaching efficiency of \EICII and \EICIII are very similar, though it might be noted that the standard deviation of \EICIII is reduced compared to \EICII, indicating the rule set did offer some assistance to the teacher. This said, the effect of providing the ruleset for the purpose of modfying the teacher's prior beliefs of the learner, $\Hfact$, is not so apparent.

These findings show how the proposed framework can be used to design an evaluation scheme for the teacher in a given task, as well as tools which are designed to support  the teacher's mental representations of the learner to improve teaching performance. See \sref{findings} for further discussion.
 
 \begin{figure}[t!]
 	
 	\begin{subfigure}[t]{\columnwidth}
 		\includegraphics[width=\columnwidth]{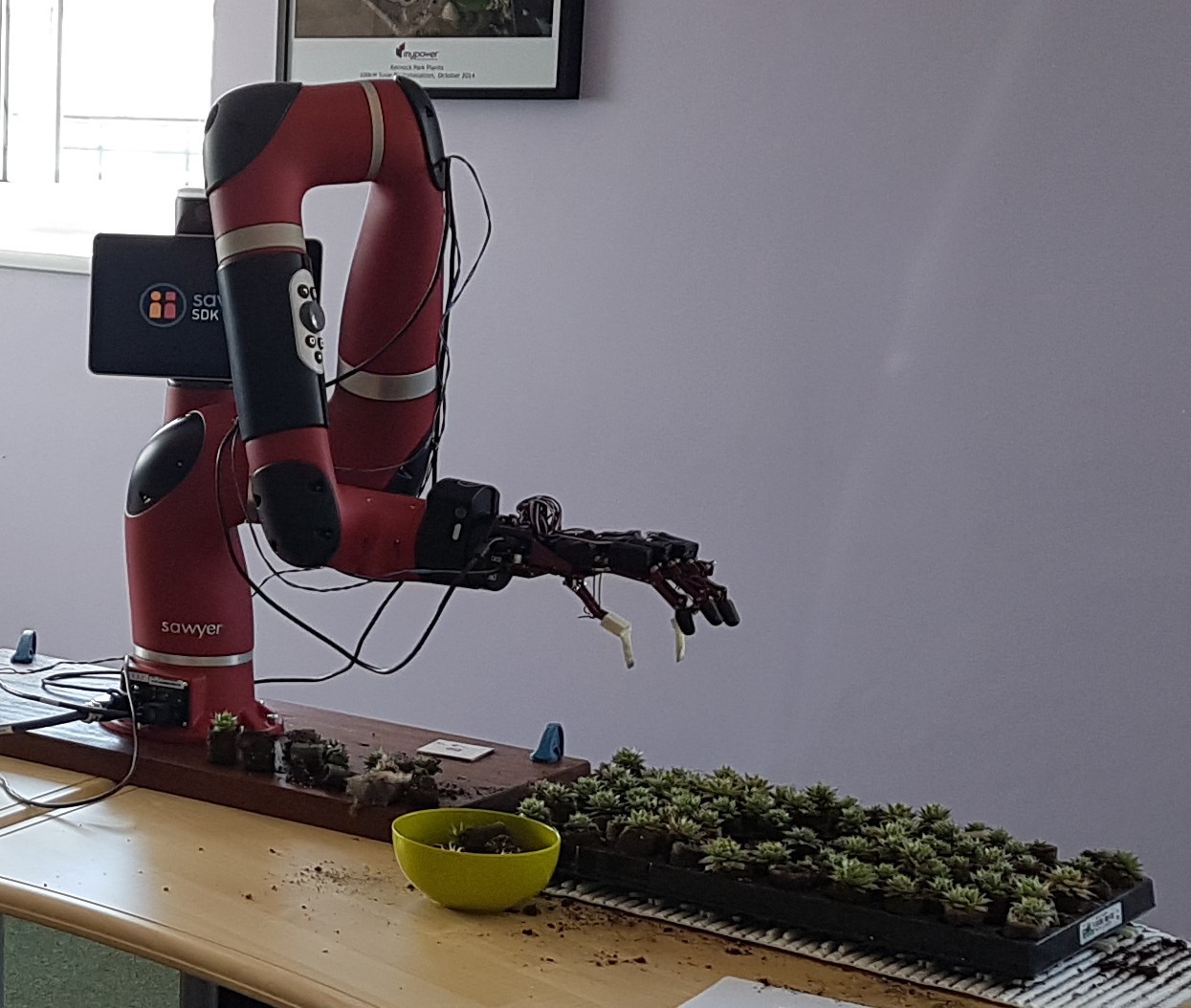}
 		\vspace{0.1cm}
	 	\put(121,0){(a)}
 	\end{subfigure}
 	\begin{subfigure}[t]{\columnwidth}
 		\begin{overpic}[width=\columnwidth]{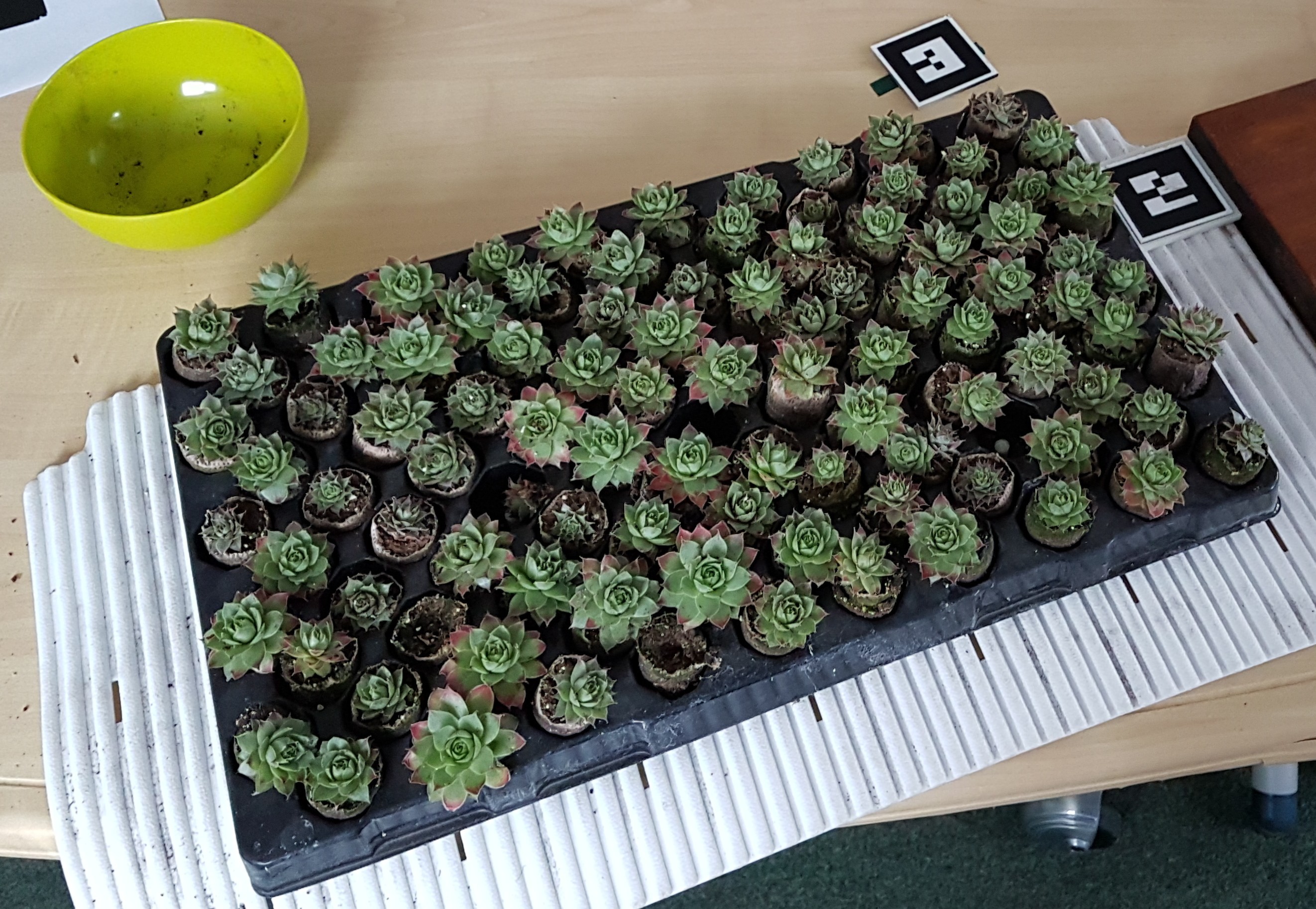}
 			\thicklines
 		{\color{red}
 		\put(13,43){\oval(5,5){}} 
 		\put(18,11){\oval(5,5){}} 
 		\put(92,34){\oval(5,5){}} 
 		\put(74,60){\oval(5,5){}} 
 		\put(53,40){\oval(5,5){}} 
 		}
		\end{overpic}
		\vspace{0.1cm}
			 	\put(121,0){(b)}
 	\end{subfigure}
 			\vspace{-0.3cm}
 	\caption{Experimental setup. (a) shows robot is shown in starting position. (b) shows the fiducial markers used to localise the plant tray and disposal bin (green bowl). Red circles in the lower image indicate the \emph{generalisation sampling} test set (see main text).}
 	\label{f:exp2_setup}
 \end{figure}
 \begin{figure*}[!]
 	\def\svgwidth{\linewidth}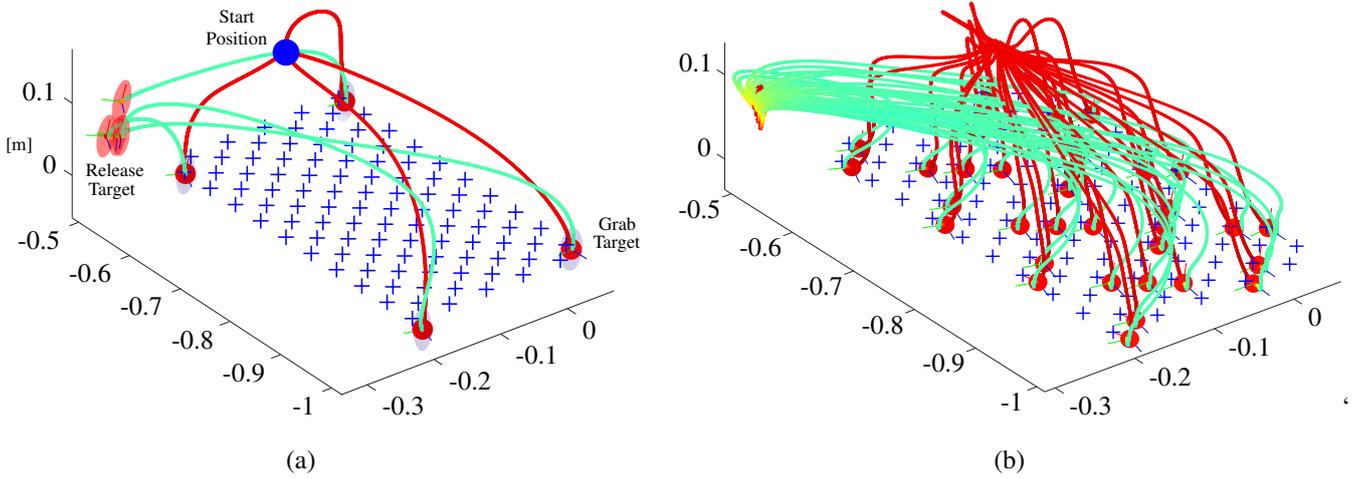
 	\put(-395,-20){(a)}
	\put(-130,-20){(b)}
	 		{\color{blue}
	 			\put(-395,137){\circle*{10}} 
	 		}
	{\scriptsize
	\put(-420,148){Start}
	\put(-425,140){Position}
	\put(-278,70){Grab}
	\put(-280,63){Target}
	\put(-470,90){Release}
	\put(-469,83){Target}
	\put(-500,100){[m]}
	}`
	\vspace{-0.1cm}
 	\manuallabel{f:demos_repos:a}{(a)}
 	\manuallabel{f:demos_repos:b}{(b)}
 	\caption{The user provides a set of demonstrations \ref{f:demos_repos:a}, which are used to build a model that is then used to generate trajectories \ref{f:demos_repos:b} for all of the targets (blue crosses). The colour of the trajectory indicates the status of the gripper, with red indicating open and green indicating closed. A subset of the generated trajectories is shown in (b) for clarity.}
 	\label{f:demos_repos}
 \end{figure*}
\subsection{Pick-and-Place Experiment}\label{s:expt-pnp}
The aim of the second experiment is to evaluate the proposed framework in a real-world scenario, where there are a large number of possible task conditions that may be encountered (\ie large $\spaceB$ and $\spaceR$). Typically, this means simple visualisation tools (such as that described in \sref{expt-ptp}) are no longer feasible, and alternative approaches must be used. An alternative to visualising task execution in such situations is to have the teacher watch the robot \emph{physically perform the skill} (\ie using a task-realisation set $\spacer$). For the teacher, this presents an additional challenge as they must now rely on a very small set of realisations $\spacer$ to form their estimates of $\spaceR$. In such a scenario, it is unclear \il{\item how selection of $\spacer$ affects novice teachers' understanding of learner abilities $\espaceR$, and \item whether they may benefit from system-selected (as opposed to self-selected) $\spacer$.} 

This experiment explores these questions through the task of plant tray sorting (see \fref{gex}), whereby the robot must be taught to reach from a start location to a target plant, grasp it, and remove it to a bin (see \fref{exp2_setup}). Note that the high number of degrees of freedom of the robot, and the number of target plants involved make $\spaceB$ and $\spaceR$ non-trivial. Additionally, as one realisation (\ie the picking and removal of a plant from a specific location) takes approximately $25$ seconds in the system considered, and there are $100$ plants in the tray, feedback to the teacher can be extremely costly in time. This makes the issue of teaching efficiency particularly important for practical deployment of such systems.

\subsubsection{Hypotheses ---}\label{s:hypotheses-2}
The experimental hypotheses are chosen to test whether teacher skill, as measured by the metrics \eref{efficacy} and \eref{teach_eff}, are influenced by the choice of feedback in form of the task realisation set $\spacer$, and whether the teachers themselves are capable of selecting informative realisations. In forming the set $\spacer$, there is flexibility in \il{\item the order in which demonstrations and realisations are interleaved, \item the specific choice of realisations in $\spaceR$ selected to form $\spacer$, and \item the responsibility of the teacher to self-select $\spacer$, or otherwise}.

To examine these issues, the following hypotheses are defined:
\begin{hypotheses}[start=4]
\item\label{E2H1} \textit{Feedback through sampling the learnt model improves teaching efficiency, compared to giving no feedback to the teacher.}
\item\label{E2H2} \textit{Generalisation sampling \emph{(see below)} of the learnt model improves teaching efficiency, compared to task realisations selected by the teacher.}
\item\label{E2H3} \textit{Generalisation sampling of the learnt model improves teaching efficiency, compared to the learner simply repeating what it was last shown.}
\item\label{E2H4} \textit{Teaching efficiency improves between episodes as the participant gains understanding of the learning process.}
\end{hypotheses}
In this experiment, for \ref{E2H2} and \ref{E2H3}, the so-called \emph{generalisation sampling} set consists of task realisations for picking the plants located in the four corners of the tray, and one plant located in its centre, see \fref{exp2_setup}. Sampling the learner's skill in these locations gives an estimate of the generalisation capabilities of the robot, since they test its behaviour across the whole space of possible target plants.

\ref{E2H2}, therefore, tests a teacher's ability to interpret feedback which has been designed to test the learner effectively,  \ie $\spacer\sim\aspaceR$.

\ref{E2H3} considers the effect of feedback which simply executes the task in the same conditions as shown by the teacher, \ie $\spacer=\spaceb$. 

\ref{E2H1} considers whether the teacher can effectively teach the learner with no feedback, $\spacer=\emptyset$.

\ref{E2H4} considers whether the feedback acts as a training mechanism in any of the conditions, \ie whether it modifies $\Hfact$ in each interaction.

\subsubsection{Materials and Methods ---}\label{s:setup-2}
Participants in this experiment are asked to teach a robot to grasp plants from a plant tray and dispose of them in a disposal bin (see \fref{exp2_setup}).

More specifically, the skill to be taught is to generate a path which, beginning from a fixed start position, must move along an obstacle-free path toward a specific target plant, into a suitable grasping position (identified by a grasping action being within a threshold distance from the plant). Once there, the robot must grab the plant and move along an obstacle-free path to a bin by the tray. Once there, it must release the plant into the bin (identified as release occurring within a threshold distance from the bin). At all stages of the movement, the gripper must remain in an admissable space $\spaceX$, which excludes self-collisions and collisions with the table. Examples of good demonstration trajectories are shown in \fref{demos_repos}~\ref{f:demos_repos:a}.

These requirements can be formally summarised as
\newcommand{\Hquad}{\hspace{-0.5em}} 
\newcommand{\Squad}{\hspace{.5em}} 
\begin{equation}
\Hquad \traj \in \spaceB \Squad \textrm{if} \Squad
\begin{cases}
(i) \Hquad &\traj \subset \spaceX \\
(ii) \Hquad &d(\traj(a_{g}), \mathbf{b}_{g}) \leq \delta_g \\
(iii) \Hquad &d(\traj(a_{r}), \mathbf{b}_{r}) \leq \delta_r
\end{cases}
,
\label{e:exp2criteria}
\end{equation}
where $d(\cdot,\cdot)$ denotes the Euclidean distance between two points, $\traj_m(a_{g})$ ($\traj_m(a_{r})$) is taken as the location of robot end-effector at the grabbing (releasing, respectively) action step, $\mathbf{b}_g$ ($\mathbf{b}_r$) are the grabbing (releasing) target locations, and $\delta_g$ ($\delta_r$) is the grabbing (releasing) threshold. The latter thresholds are defined as the mean grab distance observed in the demonstration set, $\pm 2$ standard deviations plus a $1mm$ regularisation term.

In this experiment, the start and end points of the task are fixed positions, and so the teaching must focus on achieving \emph{generalisation over the plant positions in the tray}, which shall form the test set. The test set is again constrained to a finite size, naturally discretised into a set of $100$ in the tray grid. Therefore, for this experiment $|\spaceB|=100$ and $\max |\spaceR|=100$.

As the robot learner, a Rethink Robotics Sawyer robot arm is used, equipped with an Active Robots AR10 hand. The robot arm has 7 degrees of freedom and allows for kinesthetic teaching through gravity-compensated control. The hand has a further 10 degrees of freedom, however, only four of these are used for the grab/release actions. The latter are implemented as a pincer movement (either from the open to closed position, or vice versa), triggered on-demand by the user through press of a button. Detection of the location the plant tray and disposal bin is achieved through fiducial markers \cite{Niekum2012Ar_track_alvar}, using the in-built cameras of the robot. 

During demonstrations, the teacher guides the robot through the required motions for the task using kinesthetic teaching, and the end-effector positions and orientations are recorded through the joint encoders using forward kinematics, with a repeatability of $\pm 1$mm. The teacher issues gripper control signals (open/close) using a button on the robot's end-effector cuff, and these are also recorded.

For learning, a \TPGMM model is used to encode the recorded task demonstrations (see \sref{expt-ptp}). The state representation consists of $\statept_\nd = (t_\nd, \mathbf{x}_\nd^p, \mathbf{x}_\nd^q, x^h_\nd)^\top$, where $t_\nd$ is the time stamp, $\mathbf{x}_\nd^p$ is the position of the robot's end-effector, $\mathbf{x}_\nd^q$ its orientation and $x^h_\nd$ is the hand state (either open or closed). The model is parametrised with three frames of reference: a start frame indicating the initial position of the end-effector, a target frame indicating the position of the target plant, and a goal frame indicating the position of the disposal bin. It uses a mixture of $K=7$ components, selected based on empirical testing. Examples of task realisations generated by this model can be seen in \fref{demos_repos}~\ref{f:demos_repos:b}.

\subsubsection{Procedure ---}\label{s:procedure-2}
\begin{figure*}[t]
	\small
	\def\svgwidth{\linewidth}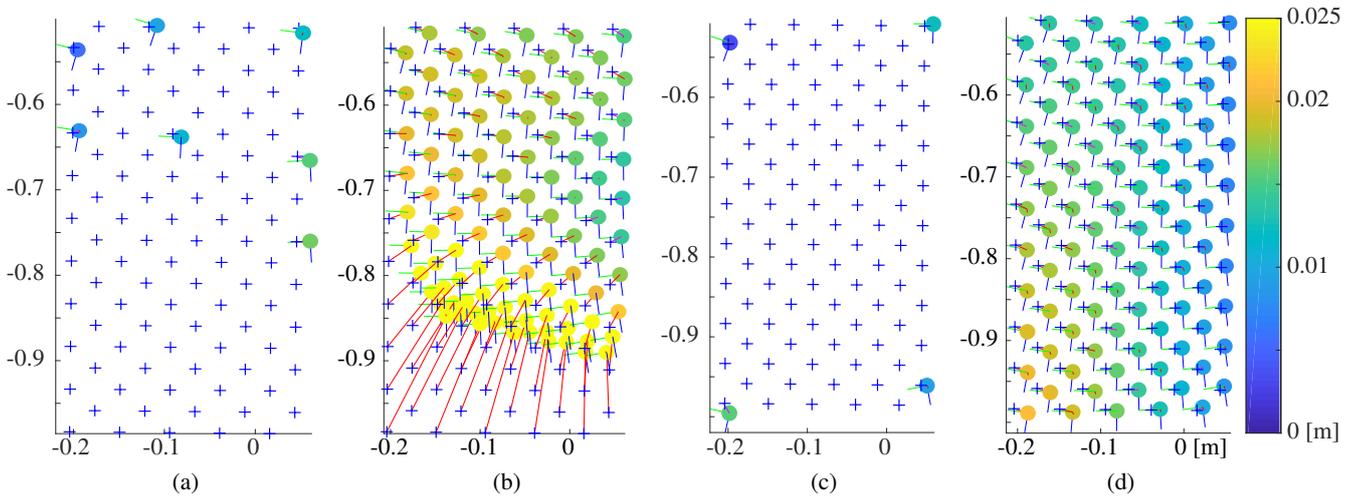
	\caption{Example of a demonstration and reproduction sets for a case of (a) a poor demonstration set and the resulting model-generated trajectories (b). (c) then shows a good demonstration set and the resulting model-generated trajectories (d). The blue crosses indicate the tray targets, the red lines indicate the intended target for a grab action, and the shaded circles indicate the location of the grab action for the given target, with the colour representing the distance of the robot grasp from the target at the time of the grab action.}
	\label{f:good_bad}
\end{figure*}

The following describes the protocol for working with the experimental participants.

Participants are screened to ensure they have no prior background in robotics or machine learning. After giving informed consent, they are assigned to one of the four interaction conditions and shown video instructions on their teaching task. Each group begins with the same introduction video, explaining that their goal is to teach the robot to perform a pick-and-place task, and provides a basic explanation of \LfD. Following this, to control for any effects relating to the participant's physical coordination skills when interacting with the robot, the participants are allowed a $5$ minute familiarisation period. During this time, they have a print-out guidance sheet showing what the buttons on the robot do\footnote{See supplementary material \textit{``Robot Controls Guidance Sheet''}.}, and are told to try pick-up as many plants from the tray as they can in the time given. 

After the familiarisation period, participants are shown a video which explains the teaching procedure and that they will be asked to teach the robot the same task three times. In each teaching session, the participant may provide as many demonstrations as they wish, but are limited to $15$ minutes per session. Participants are then shown condition-specific instructions.

In all conditions, prior to providing a demonstration the grab target must be set for the learner. The participant first points to the plant they wish to interact with and the experimenter specifies the target to the robot accordingly on a privately viewed interface (this is done to remove any possibility of the participant being affected by the system software interface). The participant then provides a demonstration of removal of the target plant into the disposal bin. This continues until they decide they have provided enough demonstrations, or the time limit for the session expires. After this point, the robot resets for the next teaching session.

The resultant data is analysed according to a mixed-factors ANOVA. A power analysis indicates that, for a medium effect size (Cohen's $f=0.3$) and a type I error rate $\alpha=0.05$, the required sample size is $\Ns=36$, or $9$ per condition \cite{Faul2009StatisticalAnalyses}. Accordingly, the results reported below are for a population of $36$ participants ($18$ male, $18$ female; ages $\mu=36.7, \sigma=10.2$). The latter were recruited from a horticultural production site (domain experts in the task of plant sorting) with all experiments conducted on-site and in a private area\footnote{This experiment was conducted with ethical approval granted by KCL REC Committee under LRS-17/18-5549.}.

\subsubsection{Conditions --- }\label{s:conditions-2} The following describes specific details of each test condition used in the experiment. 

\textit{Condition \ref{tp:E2C1} - No Feedback (NF)}\manuallabel{tp:E2C1}{1}
In this condition, participants are only able to observe the robot's task performance after they have provided a complete set of demonstrations (\ie at the end of a teaching session). No feedback is given during teaching (\ie $\spacer=\emptyset$), representing a situation where the user must rely on their own understanding of the learner's ability, $\espaceR$, developed exclusively through their knowledge of the demonstrations they provided, $\spaceb$, and their prior expectations, $\Hfact$. After a teaching session, they are permitted to view task realisations for self-selected target plants. This condition represents a typical \naive approach to \LfD, where a person provides some demonstrations of a task, and then observes the robot performing the task.

\textit{Condition \ref{tp:E2C2} - Replay Feedback (RF)}\manuallabel{tp:E2C2}{2}
In this condition, participants are shown a task realisation corresponding to the last demonstration, immediately after it is shown (\ie during teaching, after each demonstration). Here, the participant forms $\espaceR$ through observation of $\spacer$ and $\spaceb$, however, as $\spacer=\spaceb$, the feedback gives no indication of generalisation, which the participant must estimate for themselves. 

\textit{Condition \ref{tp:E2C3} - Batch feedback (BF)}\manuallabel{tp:E2C3}{3}
In this condition, task realisations for a set of pre-selected test points (\ie the generalisation sampling set) are shown to the participant after every demonstration. As noted in \sref{hypotheses-2}, these points are selected to give a good approximation of the robot's current ability $\aspaceR\sim\spaceR$. If interpreted by the participant correctly, this feedback gives information as to the extent to which generalisation has occured when forming $\espaceR$.

\textit{Condition \ref{tp:E2C4} - Selected Feedback (SF)}\manuallabel{tp:E2C4}{4}
In this condition, participants are free to choose when the robot should provide a task realisation, and under what configuration (\ie which plant location to test) during teaching. Here, the participant forms $\espaceR$ through their own, self-selected $\spacer$. 

\begin{figure}[t]%
	\centering%
	\def\svgwidth{0.8\linewidth}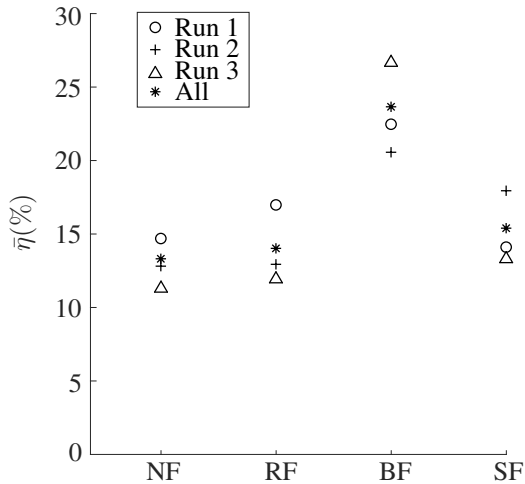
	\caption{Teaching Efficiency means ($\bar{\eta}$) for inidividual runs as well as the overall mean for the four experimental conditions.}
	\label{f:E2results}  
\end{figure}

\subsubsection{Results ---}\label{s:results-2}
The data were checked to be compatible with model assumptions. Data in \EIICI, \EIICII, and \EIICIII was found to be non-normal using an Anderson-Darling test. ANOVA tests are noted as being robust to violations of normality \cite{Schmider2010IsRobust}, and considering the excess Kurtosis for the four groups is $1.3695$, $1.6042$, $-0.6387$, and $0.9762$, respectively (where a normal distribution would have an excess kurtosis value of zero), the violation is considered minor and the data are assumed to follow a normal distribution. Using Mauchly's test for sphericity on the repeated measures model gives $\chi^2(2)=1.2423, p = 0.5372$, indicating the sphericity assumption is not violated and no data correction is required.

\begin{figure}[t!]%
	\centering%
	\def\svgwidth{\linewidth}
\begingroup%
  \makeatletter%
  \providecommand\color[2][]{%
    \errmessage{(Inkscape) Color is used for the text in Inkscape, but the package 'color.sty' is not loaded}%
    \renewcommand\color[2][]{}%
  }%
  \providecommand\transparent[1]{%
    \errmessage{(Inkscape) Transparency is used (non-zero) for the text in Inkscape, but the package 'transparent.sty' is not loaded}%
    \renewcommand\transparent[1]{}%
  }%
  \providecommand\rotatebox[2]{#2}%
  \newcommand*\fsize{\dimexpr\f@size pt\relax}%
  \newcommand*\lineheight[1]{\fontsize{\fsize}{#1\fsize}\selectfont}%
  \ifx\svgwidth\undefined%
    \setlength{\unitlength}{542.08552516bp}%
    \ifx\svgscale\undefined%
      \relax%
    \else%
      \setlength{\unitlength}{\unitlength * \real{\svgscale}}%
    \fi%
  \else%
    \setlength{\unitlength}{\svgwidth}%
  \fi%
  \global\let\svgwidth\undefined%
  \global\let\svgscale\undefined%
  \makeatother%
  \begin{picture}(1,0.61486981)%
    \lineheight{1}%
    \setlength\tabcolsep{0pt}%
    \put(0,0){\includegraphics[width=\unitlength,page=1]{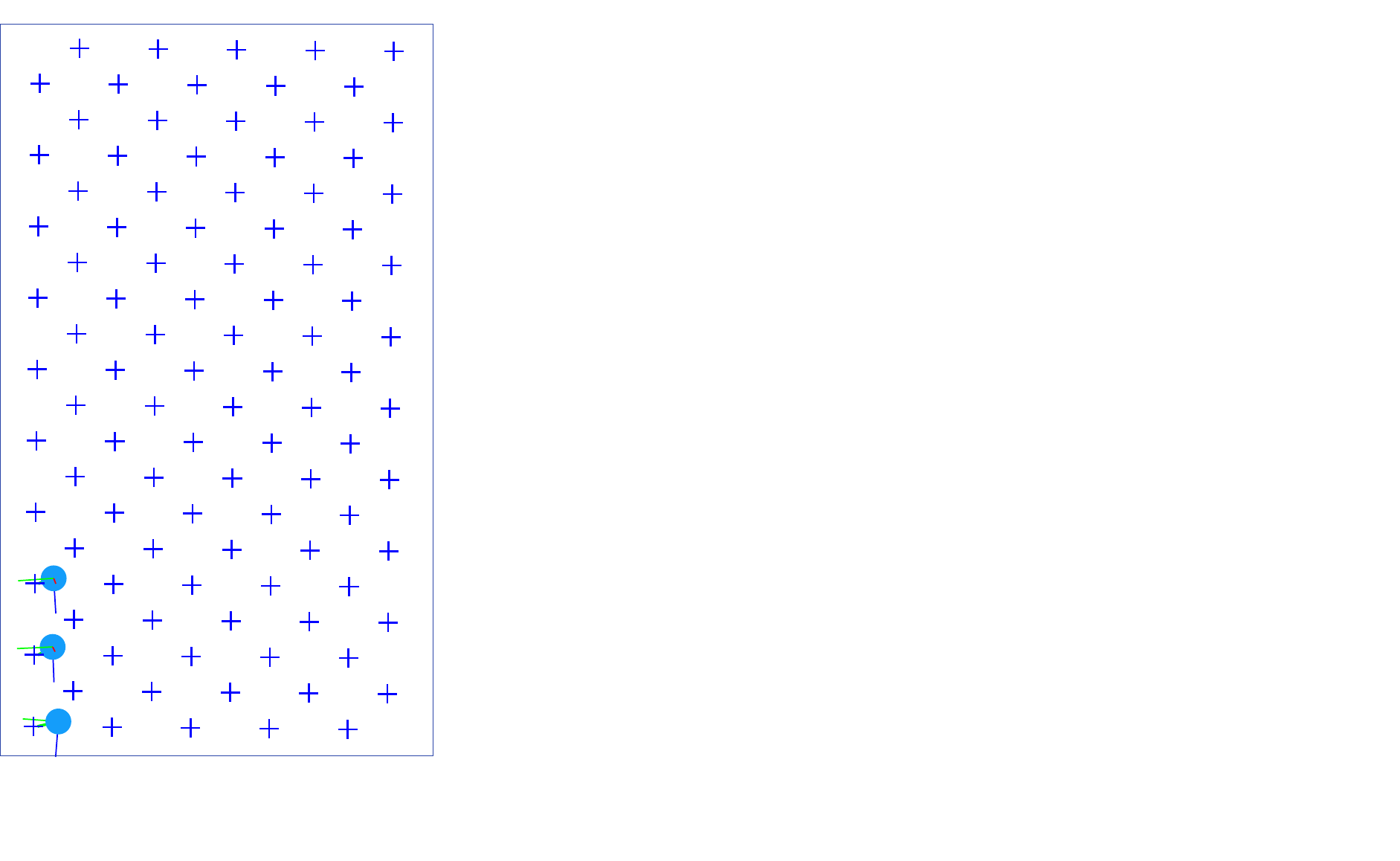}}%
    \put(0.15492659,0.00243201){\color[rgb]{0,0,0}\makebox(0,0)[t]{\lineheight{1.25}\smash{\begin{tabular}[t]{c}(a)\end{tabular}}}}%
    \put(0.49901503,0.00243201){\color[rgb]{0,0,0}\makebox(0,0)[t]{\lineheight{1.25}\smash{\begin{tabular}[t]{c}(b)\end{tabular}}}}%
    \put(0,0){\includegraphics[width=\unitlength,page=2]{failures.pdf}}%
    \put(0.84508242,0.00243201){\color[rgb]{0,0,0}\makebox(0,0)[t]{\lineheight{1.25}\smash{\begin{tabular}[t]{c}(c)\end{tabular}}}}%
    \put(0,0){\includegraphics[width=\unitlength,page=3]{failures.pdf}}%
  \end{picture}%
\endgroup%

	\manuallabel{f:failures:a}{(a)}
	\manuallabel{f:failures:b}{(b)}
	\manuallabel{f:failures:c}{(c)}
	\vspace{-0.5cm}
	\caption{Examples of failure modes observed in demonstration sets. \ref{f:failures:a} undemonstrated states (demonstrations do not cover the space), \ref{f:failures:b} ambiguous demonstrations (multiple demonstrations co-located at top of grid), \ref{f:failures:c} errors made during demonstration (demonstration given for wrong target: the red line links the desired target with the demonstration actually given).}
	\label{f:failures}      
\end{figure}

Considering the teaching efficiency for the participant groups, there was a significant main effect observed for the between-subjects factor, indicating a difference was observed between the test conditions (\EIICI, \EIICII, \etc), $F(3,32) = 13.864, p = 5.797\times10^{-6}$. As significance was found in the main effect, a multiple comparison of means was performed. This shows that \EIICIII provides a $10.76\%$ improvement in efficiency compared to \EIICI ($p=1.017\times10^{-5}$), 
$9.64\%$ compared to \EIICII ($p=6.2947\times10^{-5}$), and $8.20\%$ compared to \EIICIV ($p=5.8708\times10^{-4}$), see \fref{E2results}. 

There was no significance found in the within-subjects factor analysis, \ie there was no support for a learning effect improving performance over each consecutive run over all conditions. However, an interaction effect was observed for the between and within factors, $F(6,64)=2.7845, p=0.01808)$. As an interaction effect was found, a multiple comparison of means conditioned on the within factor is used to consider the \textit{simple main effects}. This multiple comparison indicates that for the first run, no significance in improvement is observed between \EIICIII and \EIICII, and no significance is observed between \EIICIII and \EIICIV. On the third run, significance is observed between \EIICIII ($27.04\%$) and each of the other conditions. In this third run, with \EIICI ($11.35\%$), \EIICII ($11.94\%$), and \EIICIV ($13.7\%$), the performance of \EIICIII represents an average improvement of $\sim220\%$ relative to each condition. On average, across all runs, the mean teaching efficiency of \EIICIII represents a $\sim169\%$ relative improvement over the other conditions.

To gain further insight into the participants' teaching behaviour, the occurance of \il{\item demonstration errors, \item ambiguous demonstrations, and \item undemonstrated states} are recorded in \tref{error_table}. The demonstration error values in \tref{error_table} are gathered by visual inspection of the demonstration sets (see \fref{failures}(c)), while the values for undemonstrated states and ambiguous demonstrations are generated by post-processing the data for each participant iteratively to see the performance of the learnt model after each demonstration is provided, and then applying the definitions of ambiguous demonstrations \eref{ambig_alt} and undemonstrated states data sets \eref{undemonstrated-e2}. Typical examples of these failures are shown in \fref{failures}.

Looking at the mean performance levels achieved across all runs for each condition, it can be seen that \EIICIII achieved the best overall performance, with $23.54\%$ teaching efficiency. This represents an average improvement of $\sim169\%$ relative to \EIICI ($12.78\%$), \EIICI ($13.89\%$), and \EIICIV ($ 15.34\%$). The maximum average teaching performance can be found on the third run, where \EIICIII achieved $27.04\%$, representing  an average improvement of $\sim220\%$, relative to \EIICI ($11.35\%$), \EIICII ($11.94\%$), and \EIICIV ($13.7\%$).

From these results and the statistical analysis, there is support for \ref{E2H2}, \ref{E2H3}, but not \ref{E2H1} or \ref{E2H4}. See \sref{findings} for further discussion.

\begin{table}[H]
	\centering
	\caption{Total number of error occurrences observed in participant data for each condition. In this experiment, demonstration errors occur when participants provided an example for a plant different to the one they selected. Undemonstrated state counts are calculated using the cardinality of the set defined by \eref{undemonstrated-e2}, and ambiguous demonstrations are found by considering the learner's efficiacy as defined by \eref{ambig_alt}.}
	\small
	\begin{tabularx}{0.9\columnwidth}{rcccc}
		\toprule
		~ & \textbf{\EIICI} & \textbf{\EIICII} & \textbf{\EIICIII} & \textbf{\EIICIV} \\ \midrule
		Demonstration Errors & 4             & 0             & 3             & 3             \\ \midrule
		Undemonstrated States   & 499             & 619             & 827             & 485             \\ \midrule
		Ambiguous Demonstrations   & 54             & 51             & 6             & 39             \\ \midrule
	\end{tabularx}
	\label{t:error_table}
\end{table}

\section{Discussion}\label{s:findings}
The results from both experiments show strong evidence that the feedback systems used could significantly improve participants' teaching, resulting in improved learner performance on the target tasks. By using the proposed framework to design evaluation and feedback focused on their needs (their interpretation of $\spacer$, $\espaceR$, \etc), they develop an understanding of what demonstrations the learner \textit{requires} to learn, without necessarily understanding \textit{how} learning is taking place.

In addition to the basic result of improving teaching performance, a number of insights can be gained from the experimental results. In the first experiment, it can be seen that unguided participants (\ie those in the \EICI condition) tend to underestimate the number of demonstrations required to teach effectively, compared to the other conditions (see \fref{icra_results2}\ref{f:icra_results2:a}). This is an indicator that the participants' prior beliefs about the learner, $\Hfact$, did not match reality, resulting in a poor estimate of the learner's ability ($\espaceR\neq\aspaceR$). Looking at the efficacy (see \fref{icra_results2}\ref{f:icra_results2:b}) it is seen that feedback in the \EICII and \EICIII cases not only improves performance (higher mean efficacy), but leads to greater consistency (lower standard deviation) among participants,  compared to \EICI. This highlights the difficulty unguided teachers had in providing adequate demonstrations. While some ($7$ participants) did manage to achieve reasonably good performance in the \EICI condition, there was wide variation, and the majority performed badly. Conversely, in \EICII and \EICIII, \textit{all} participants reached close to the maximum efficacy, indicating that the visualisation helped shape their understanding of the teaching task ($\espaceB$ and $\espaceR$), and thereby enabled them to provide better demonstrations.

The second experiment also supports these findings, and gives further insight into \il{\item what kinds of feedback are beneficial, and \item the common pitfalls novice teachers encounter}. Looking at \fref{E2results}, it can be seen that there is a significant boost to performance observed in the \EIICIII condition, but feedback in the \EIICII and \EIICIV conditions, offer little improvement over the \EIICI case. The results indicate that for feedback to be of benefit to the teacher, it must offer insight on \emph{how well the learner generalises}, \ie simply showing the teacher replays of demonstration conditions, $\spacer=\spaceb$ as in the \EIICII case, is not sufficient guidance for teaching. In addition, novice users cannot be expected to know how to test for generalisation without appropriate training, as indicated in the \EIICIV condition.

To gain insights into the reasons for this, it is useful to look at participant teaching behaviours. Looking at \tref{error_table}, \EIICIII featured the fewest ambiguous demonstrations, but also the most undemonstrated states on average. This suggests that the demonstrations provided by participants under the \EIICIII condition are nearly always useful for the learning system, however, the participants are not providing complete demonstration sets. This may either be due to the participants deciding the system has been shown enough information on the task and stopping prematurely (poor $\espaceR$), or they are not able to provide enough demonstrations in the $15$ minute period permitted (feedback in this condition is the most costly in time, with $5$ task realisations shown to the teacher after every demonstration). Nevertheless, performance is highest in this condition, suggesting the teaching to be of higher quality.

In addition to these observations, while \EIICII did not provide good overall performance, some benefits to this condition can be seen. Looking at \tref{error_table}, participants made no demonstration errors in this condition. Having the robot replay what the participants had just demonstrated appears to help them spot errors as they happen (\eg providing a demonstration for the wrong target). This suggests that it may be beneficial to design elements of the \EIICII and \EIICIII strategies that combine feedback both on immediate errors and learner generalisation.

\section{Summary}
This paper presents an extended model for \LfD, which incorporates the teacher's understanding of, and influence on, the learner. The proposed framework introduces a new space---the teacher's belief space---to the standard view on \LfD, highlighting the teacher's understanding of the learner's ability, and how this forms the basis of their interaction. The ability to formalise this relationship, and develop quantitative metrics for its study, is crucial given that learner performance strongly depends on the quality of data provided by the teacher.

The two experiments reported here show the benefit of approaching \LfD problems using the proposed framework, by enabling measures for assessing teaching quality, and identifying teaching failures, to be defined. These, in turn, can be used in creating feedback tools to directly influence novice teacher behaviour and guide them toward better teaching practice. Results from the experiments show that, without this guidance, novice teachers struggle to efficiently provide demonstrations which avoid issues like undemonstrated states and ambiguity, even for relatively simple teaching tasks. In contrast, using feedback designed from the perspective of improving teaching quality, as guided by the proposed model, can result in a teaching efficiency improvement of $\sim169-180\%$. These results point to the practical benefits of the proposed model, and it is hoped that this approach to incorporating teachers' thought processes more directly into \LfD will help improve novice interactions with \LfD systems.

A limitation of the presented framework is the difficulty in identifying the underlying objective task which the user wishes to achieve. Future work on modelling \LfD with consideration of the teacher would therefore include overcoming this to autonomously identify optimal test sets for general \LfD tasks which are economic with the teaching effort required.

Overall, it is hoped that the presented framework and supporting results highlight the benefit of directly modelling the teacher-robot interactions during \LfD, and the resulting new opportunities for evaluating and improving  teaching and learning of robotic grasping tasks. By providing a useful structure to \LfD problems, feedback tools can be designed to enable novice users to better leverage sophisticated policy learning methods to provide robots with advanced manipulation skills that would be otherwise difficult to train, and difficult to verify learning success.


\section{Acknowledgements}
This research was supported by the UK Agriculture and Horticulture Development Board (AHDB), under project  HNS/PO 194 - \textit{GROWBOT: A Grower-Reprogrammable Robot for Ornamental Plant Production Tasks}, and by the Engineering and Physical Sciences Research Council (EPSRC), under project EP/P010202/ - \textit{SoftSkills: Soft Robotic Skill Learning from Human Demonstration}.
\bibliography{references.bib}{}
\bibliographystyle{SageH}
\end{document}